\documentclass{article}
\usepackage{hyperref} 
 \usepackage[preprint]{neurips_2026}

\usepackage[utf8]{inputenc} 
\usepackage[T1]{fontenc}    

\usepackage{url}            
\usepackage{booktabs}       
\usepackage{amsfonts}       
\usepackage{nicefrac}       
\usepackage{microtype}      
\usepackage{xcolor}         

\usepackage{listings}
\usepackage{graphicx}
\usepackage{microtype}
\usepackage{tabularx}
\usepackage{subcaption}
\usepackage{booktabs} 
\usepackage{multirow}
\usepackage{subcaption}

\usepackage{amsmath}
\usepackage{amssymb}
\usepackage{mathtools}
\usepackage{amsthm}
\usepackage{lineno}

\definecolor{codegreen}{rgb}{0,0.6,0}
\definecolor{codegray}{rgb}{0.5,0.5,0.5}
\definecolor{codepurple}{rgb}{0.58,0,0.82}
\definecolor{backcolour}{rgb}{0.95,0.95,0.92}

\title{On the Persistent Effects of Lexicality in Large Language Models}

\author{%
  Hammad Rizwan \\
  Dalhousie University\\
  Halifax, NS\\
  \And
  Muhammad Umair Haider \\
  University of Kentucky \\
  Lexington, KY \\
  \And
  Nishant Subramani \\
  Carnegie Mellon University \\
  Pittsburgh, PA \\
  \And
   Mona T. Diab \\
  Carnegie Mellon University \\
  Pittsburgh, PA \\
  \And
  A.B. Siddique\\
University of Kentucky \\
  Lexington, KY \\
\And
  Hassan Sajjad \\
  Dalhousie University\\
  Halifax, NS\\
}

\begin{document}

\maketitle

\begin{abstract}


Representations extracted from large language models (LLMs) play an important role in many downstream applications. However, the structure of these representations is often influenced by lexical overlap rather than semantic content. Our understanding of the relationship between this lexical influence and semantic content, and its implications for downstream tasks, remains limited. In this work, we investigate representations to quantify the effect of lexical overlap relative to semantic content. We consider several adversarial semantic stress tests and further connect our findings to the information theory perspective. We find that lexical influence extends across the depth of models, consistently across architectures, training regimes, and objective functions, including the models trained for semantic similarity. Moreover, we observe a mid-depth region in which both lexical and semantic signals degrade simultaneously, indicating a transitional regime where representations are poor for both surface form and meaning. We further demonstrate the effect of lexical influence on downstream uses of LLMs using summarization and model editing as a case study. 


\end{abstract}

\section{Introduction}

Natural Language Processing (NLP) has seen rapid advances with the emergence of LLMs, which now achieve strong performance across a wide variety of benchmarks and downstream tasks. Beyond generating text, these models are increasingly used as general-purpose embedding engines, providing vector representations that underpin retrieval, textual similarity, clustering, and evaluation pipelines. 

The widespread reliance on LLM embeddings implicitly assumes that, as we progress through model depth, representations gradually move away from local lexical structure and converge toward increasingly abstract sequence-level semantics~\citep{JawaharSS19, HewittL19,haider2024looking}. In practice, this assumption is fragile, since it does not imply that lexical cues are eliminated as representations can remain highly similar whenever inputs share tokens, even when their meanings diverge~\citep{DumpalaJSMO024,RizwanRW025}. When lexical overlap influences embedding geometry, similarity estimates degrade, making it harder to retrieve or classify documents based on meaning rather than wording and weakening systems that depend on meaning-driven representations.
We therefore ask how representational geometry evolves along the axes of lexical and semantic information; how training regimes, including pretraining objectives, instruction tuning, and contrastive learning, shape these geometric structures; and how lexical influence propagates to downstream uses of LLMs.

We treat every layer of a model as a candidate embedding space and organize our study into four components. \textbf{First}, to quantify lexical influence, we use a triplet semantic-equivalence stress test consisting of an anchor, a meaning-preserving paraphrase, and a meaning-changing distractor that shares substantial lexical overlap with the anchor. This construction isolates semantic similarity failures driven by surface-form similarity and localizes the depths at which representations most often conflate lexical overlap with semantic equivalence.
\textbf{Second}, to quantify these failures in terms of lexical and semantic signals, we pair the stress tests with two layer-wise measurements: lexical decodability and semantic fidelity. Lexical decodability is quantified by training linear token-identity probes on hidden states, measuring how much surface form remains linearly recoverable at each layer. We hypothesize that a strong token identity corresponds to a salient lexical signal that downstream methods can readily leverage, whether they rely on similarity geometry or learn additional predictors. We quantify semantic fidelity based on the representation's performance on diverse embedding tasks. Together, these diagnostics distinguish among competing explanations for representation-similarity failures: lexical dominance, weak semantic organization, and transitional regimes in which both lexical and semantic signals are simultaneously low. 
\textbf{Third}, as layer-wise lexical and semantic geometry does not by itself characterize \emph{information processing dynamics}, we 
relate our investigation to these dynamics using
\emph{input (prompt) entropy} across layers. We test whether information compression and decompression shifts correlate with the layer-wise changes in lexical and semantic structure.
\textbf{Fourth}, we 
show
the 
consequences of lexical influence in practical LLM applications, focusing on abstractive summarization evaluation and factual model editing.




Our investigation leads to several central insights:

\textbf{Lexical influence persists across depth (\S~\ref{sec:measure_lex_infl}).} Adversarial stress tests across multiple models reveal that token-overlap-driven similarity errors persist throughout model depth. Although attenuated in deeper layers, the effect is neither fully eliminated nor confined to shallow representations.

\textbf{Adaptation regimes improve embeddings, but do not remove lexical influence (\S~\ref{sec:measure_lex_infl}).}  Instruction tuning and metric learning improve overall embedding quality, yet our stress tests show that metric learning still fails under lexical overlap adversaries. Lexically similar but semantically mismatched pairs continue to receive inflated similarity scores, indicating that the dominant training paradigm for embedding models reduces but does not eliminate lexical bias.

\textbf{Lexical decodability is non-monotonic across depth (\S~\ref{sec:lex_probe}).}
The ability to decode exact lexical identity from token representations fluctuates across layers rather than declining monotonically. This is in line with findings by~\cite{chengemergence},
indicating that depth does not induce a clean lexical-to-semantic transition~\citep{TenneyDP19,JawaharSS19,li2025model}.

\textbf{A mid-depth valley where lexical and semantic signals both weaken (\S~\ref{sec:lex_semantic_probes}, \S~\ref{sec:input_entropy})}. Across model families, intermediate layers form a valley-like region where token identity is least recoverable, and semantic performance stagnates or degrades across both raw embedding geometry and linearly probed semantic evaluations. This valley appears to align with a compression–re-expansion point in prompt information across layers; a similar valley shows up under full attention even when representation entropy stays roughly constant with depth, indicating that the effect is not simply an entropy transition.

\textbf{Practical implications (\S~\ref{sec:practical_implications}).}
In embedding-based pipelines, lexical overlap can miscalibrate similarity signals, degrading performance. In summarization evaluation, common reference-based metrics systematically reward reference wording, favoring surface overlap over semantic preservation. In weight-space model editing, updates generalize along surface-form similarity, producing correlated shifts for token-overlapping distractors, compromising edit locality.

\section{Experimental Setup and Preliminaries}
\label{sec:experiment_setup}
We use a common set of datasets and model families to ensure consistency across experiments.

\textbf{Datasets.} To probe lexical influence, we use two adversarial benchmarks, CounterFact~\citep{MengBAB22} and SugarCrepe++ (SCPP)~\citep{DumpalaJSMO024}. Relative to CounterFact, which primarily perturbs entities, SCPP offers a more targeted benchmark for lexical-influence tests, as it introduces systematic shifts in attributes and relations. We additionally use CounterFact for model editing. Dataset details and reference samples are provided in Appendix~\ref{sec:appendix_stress_editing_data}. To measure lexical decodability, token-identity probes are trained on WikiText~\citep{MerityX0S17}, and to evaluate semantic fidelity, the MTEB benchmark~\citep{MuennighoffTMR23} is utilized.

\textbf{Models.} We consider
three training paradigms: \emph{pretrained}, \emph{instruction-tuned}, and \emph{metric learning trained} embedding models, spanning multiple model families.\footnote{Llama~3.2~\citep{grattafiori2024llama} and Gemma~3~\citep{team2025gemma} in both \emph{pretrained} and \emph{instruction-tuned} variants, along with \emph{embedding} models including Qwen-3~\citep{qwen3embedding} (Qwen3-Embedding-8B) and KaLM~\citep{zhao2025kalmembeddingv2} (KaLM-Embedding-Gemma3-12B-2511). Note: KaLM modifies the Gemma3 architecture to use full attention.}




\section{Measuring Lexical Influence}
\label{sec:measure_lex_infl}

In the context of learned representations, we define \emph{lexical influence} as a semantic failure case in which an anchor prompt is embedded closer to a lexically overlapping distractor than to its meaning-preserving paraphrase, violating the intended triplet ordering. We measure this effect on a dataset of triplets $\mathcal{D}=\{(a,p,d)\}$, where $a$ is an anchor, $p$ a meaning-preserving paraphrase, and $d$ a lexical distractor. Dataset usage and details are provided in Appendix~\ref{sec:appendix_sample_details_stress}.

For a given model $M$ and input $x$ consisting of tokens $t \in \{1,\dots,T_x\}$ , let $H_\ell^{M}(x)\in\mathbb{R}^{T_x\times d}$ denote the matrix of token-level hidden states at layer $\ell \in \mathcal{L}$, where the $t$-th row $H_\ell^{M}(x)_t\in\mathbb{R}^d$ corresponds to token position $t$. Let $h_\ell^{M}(x)\in\mathbb{R}^{d}$ denote the corresponding sentence-level embedding, obtained via one of two standard choices: mean pooling and last token.
\[
h_{\ell}^{M,\text{mean}}(x)=\frac{1}{T_x}\sum_{t=1}^{T_x} H_\ell^{M}(x)_t,
\quad
h_{\ell}^{M,\text{last}}(x)=H_\ell^{M}(x)_{T_x}
\]
We then $\ell_2$-normalize $\tilde h_\ell^{M}(x)=h_\ell^{M}(x)/\|h_\ell^{M}(x)\|_2$ and measure 
euclidean distance,
\[
d_\ell^{M}(x,y)=\bigl\|\tilde h_\ell^{M}(x)-\tilde h_\ell^{M}(y)\bigr\|_2 .
\]
For $\ell_2$-normalized vectors, this is equivalent (up to a monotone transform) to cosine similarity, since $\bigl\|\tilde u-\tilde v\bigr\|_2^2 = 2 - 2\,\tilde u^\top \tilde v$.
To quantify lexical influence, we measure a triplet success rate. When representations are driven by lexical overlap, the model is more likely to confuse a paraphrase with a lexically overlapping distractor, yielding a lower success rate. For each triplet $(a,p,d)$, success occurs when the anchor and paraphrase are closer than either pair involving the distractor. We report the layer-wise success rate
\[
\mathrm{SR}_\ell^{M}
=
\frac{1}{|\mathcal{D}|}
\sum_{(a,p,d)\in\mathcal{D}}
\mathbb{I}\!\left[
d_\ell^{M}(a,p)
<
\min\!\left\{
d_\ell^{M}(a,d),
d_\ell^{M}(p,d)
\right\}
\right].
\label{eq:sr}
\]

\begin{figure*}[!t]
    \centering
    \begin{subfigure}{0.48\textwidth}
        \centering
        \includegraphics[width=\linewidth]{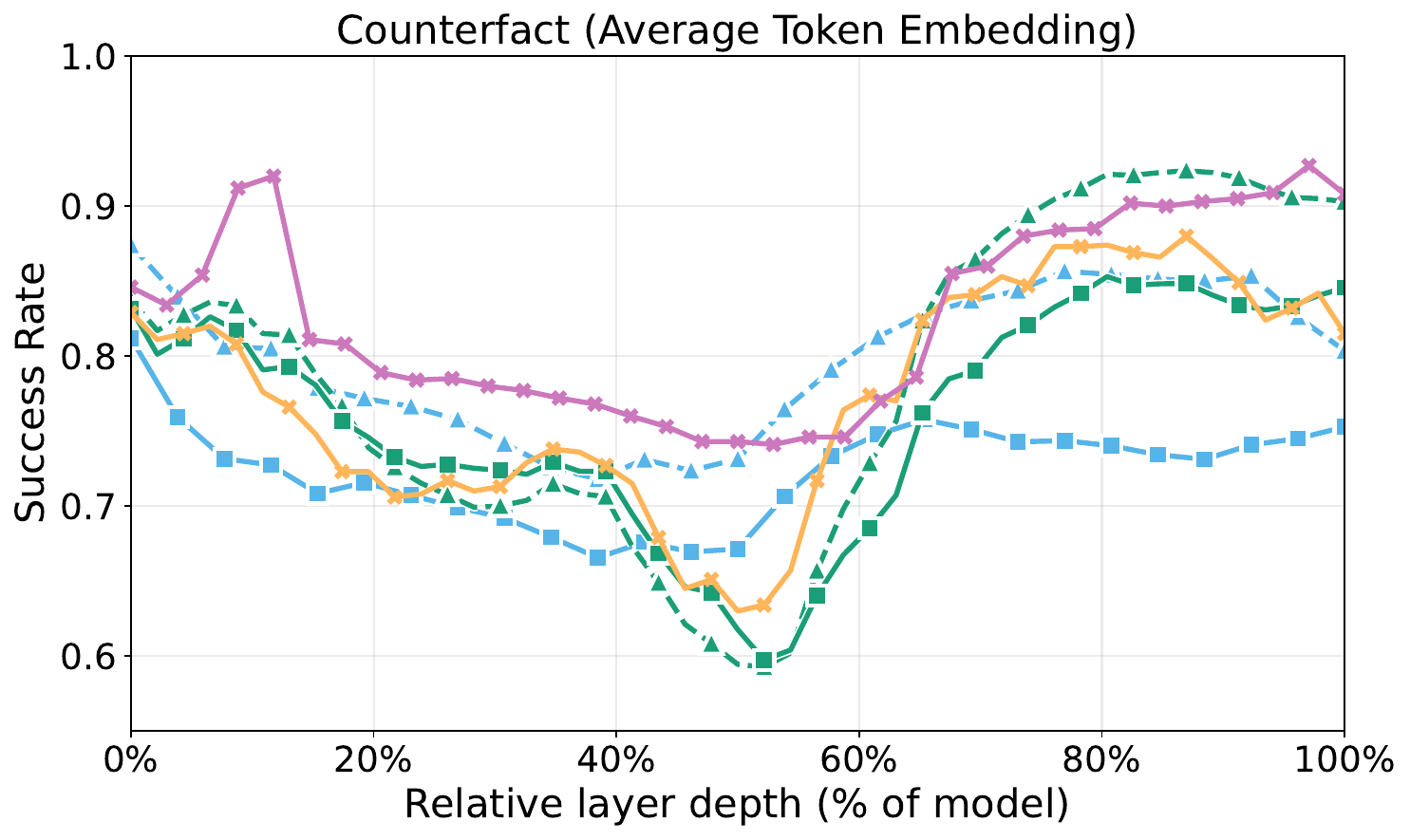}
        \caption{}
        \label{subfig:1a}
    \end{subfigure}
    \hfill
    \begin{subfigure}{0.48\textwidth}
        \centering
        \includegraphics[width=\linewidth]{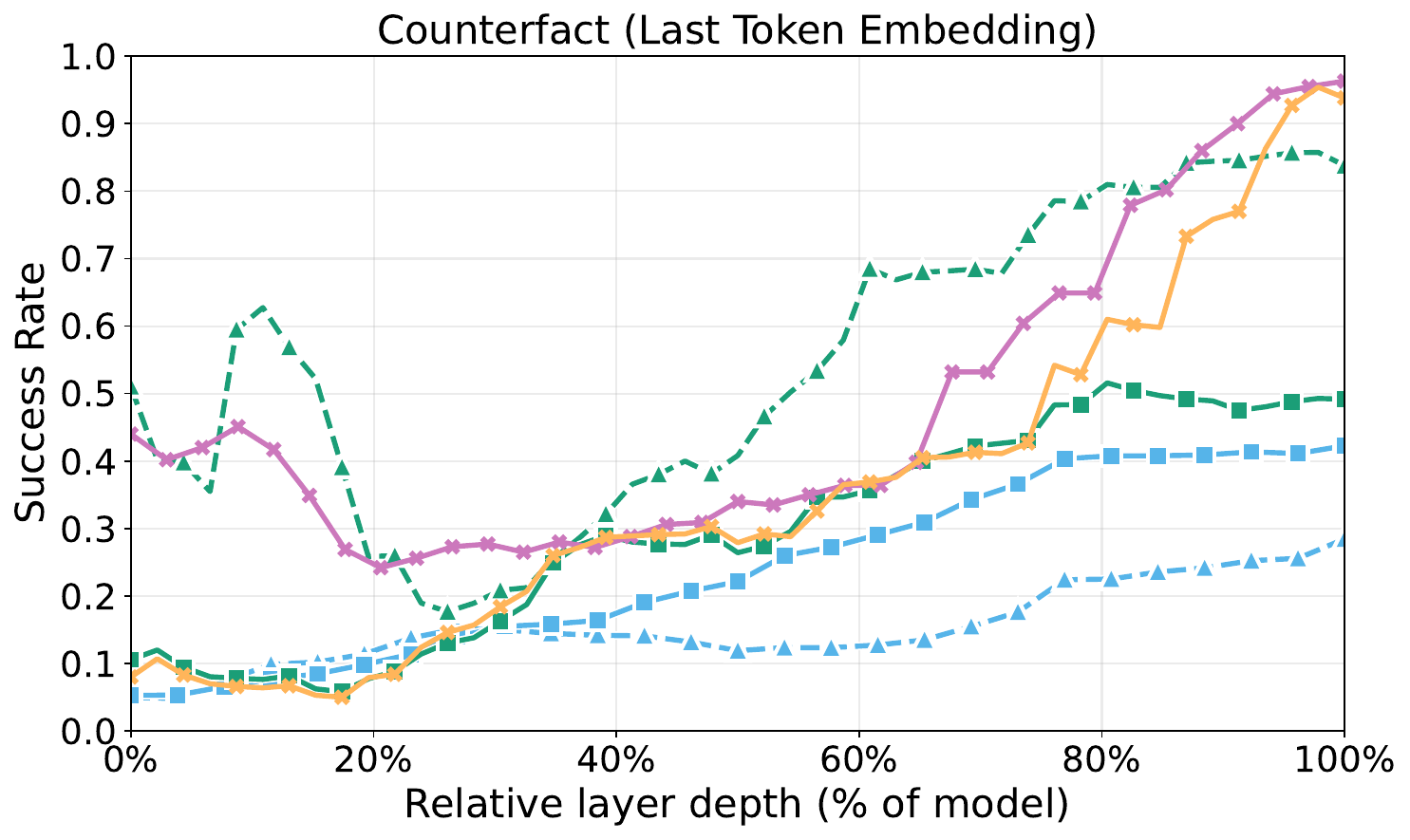}
        \caption{}
        \label{subfig:1b}
    \end{subfigure} 
    \begin{subfigure}{0.48\textwidth}
        \centering
        \includegraphics[width=\linewidth]{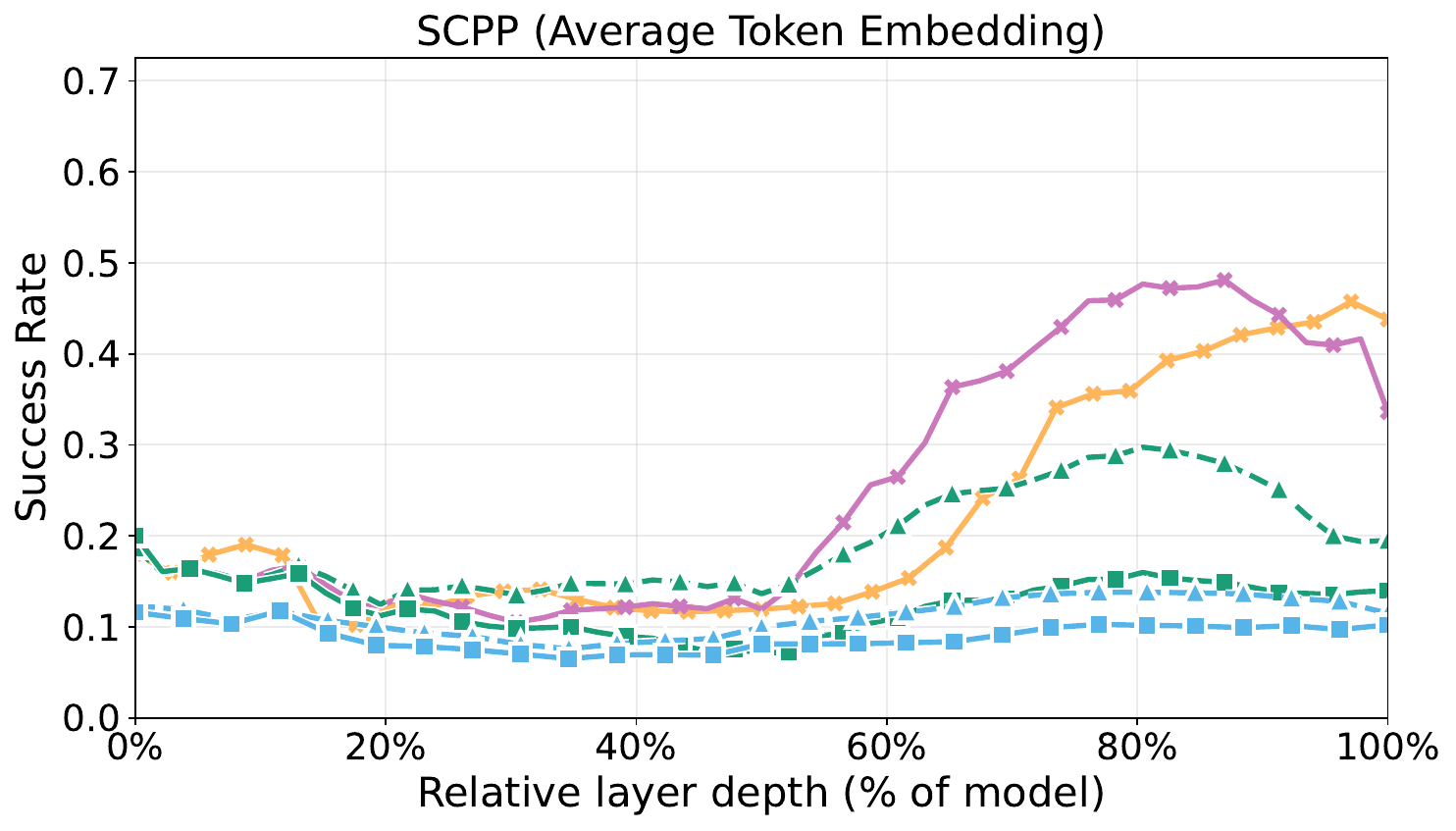}
        \caption{}
        \label{subfig:1c}

    \end{subfigure}
    \hfill
    \begin{subfigure}{0.48\textwidth}
        \centering
    \includegraphics[width=\linewidth]{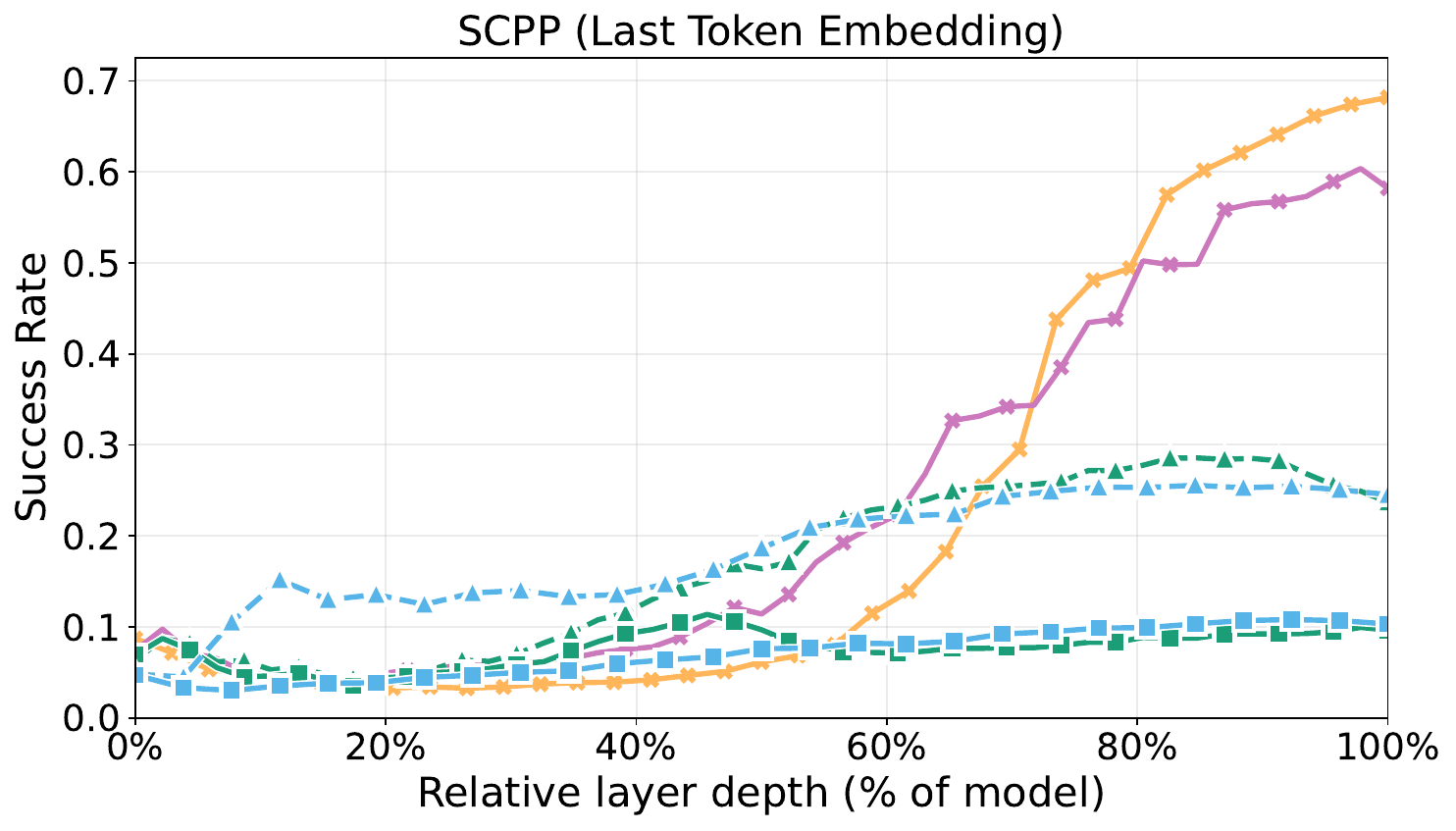}
        \caption{}
        \label{subfig:1d}
    \end{subfigure}
    \begin{subfigure}{\textwidth}
        \centering
\includegraphics[width=0.98\textwidth]{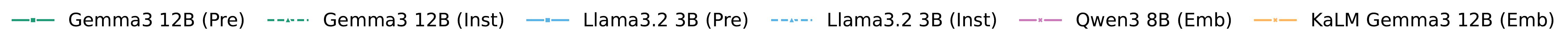}
    \end{subfigure}
    \caption{
    Success rates for lexical influence tests on CounterFact and SCPP across models using average-token and last-token embeddings. Higher success indicates lower lexical influence. Panels show CounterFact with (a) average-token and (b) last-token embeddings, and SCPP with (c) average-token and (d) last-token embeddings.
    }
    \label{fig:lexical_failures}
    \vspace{-10pt}
\end{figure*}

Figure~\ref{fig:lexical_failures} compares averaged token (mean-pooled) embeddings (\ref{subfig:1a},~\ref{subfig:1c}) and last-token embeddings (\ref{subfig:1b},~\ref{subfig:1d}) on CounterFact and SCPP (averaged results are reported for SCPP, task-wise results are provided in Appendix~\ref{sec:appendix_scpp_results}). Across pretrained and instruction-tuned models, averaged token representations generally outperform last-token embeddings; the gap is clearer on CounterFact, where last-token embeddings consistently underperform, suggesting stronger lexical influence and that single-token summaries are under-contextualized for sentence-level meaning, while on SCPP the last token is more often comparable, and the pooling advantage is typically smaller. This pattern is consistent with prior work suggesting pooling encourages more semantically abstract representations than relying on a single position~\citep{reimers2019sentence,xing2024comparative}. We also observe a consistent depth effect. Average token embeddings often exhibit a mid-layer valley, where performance briefly declines before recovering and stabilizing in the upper layers. By contrast, last-token representations typically stagnate over the same depth range, or improve only gradually. Only in the later layers do models more reliably separate true paraphrases from lexically similar distractors.


Results for \textbf{Pretrained and Instruction-tuned models}  suggest that lexical influence is strongly shaped during the core language-model pretraining phase, where the model is optimized for next-token prediction and therefore to match broad corpus statistics. In this regime, next-token prediction can make it statistically efficient for the model to rely on surface-level lexical cues (e.g., entity names, salient attributes) that correlate with the target~\citep{DuHZTH24,BandelGE22}. This helps explain why lexical-overlap failures appear in both pretrained and instruction-tuned models. Post-training can improve instruction following and response quality~\citep{Ouyang0JAWMZASR22,ChungHLZTFL00BW24}, but it operates on top of a pretrained representation space and may not fully remove lexical shortcuts inherited from pretraining~\citep{SerranoDS23}.


The layer-wise patterns reinforce this picture. Up to roughly the mid-depth of the model, the success rates of pretrained and instruction-tuned models remain closely aligned, indicating that instruction tuning does not substantially reshape the earlier layers of the representation hierarchy. Beyond this point, the success rate of instruction-tuned models increases more rapidly than that of the purely pretrained model, suggesting that later layers are selectively adapted to support more instruction-following and semantically appropriate behavior~\citep{zhao2024-layer}. However, lexical failures on SCPP remain high even in the upper layers, suggesting that lexical influences are deeply embedded in the representational structure learned during pretraining. This is consistent with evidence that fine-tuning primarily reweighs features already extractable from the pretrained model~\citep{LoveringJLP21}, and that shortcut reliance can persist in LLMs despite post-training interventions~\citep{YuanZZZL24}. These findings are robust to prompt variations, as reflected by high distance correlations between layer-wise curves for both average and last-token embeddings (0.993 and 0.78, respectively; Appendix~\ref{sec:appendix_prompt_variation}).




\textbf{Embedding models} perform near-perfectly on the CounterFact dataset; paraphrases and distractors are almost always correctly separated in the learned embedding space. This is unsurprising given that these models are trained explicitly for semantic textual similarity and the structure of the dataset. CounterFact primarily perturbs entity tokens while keeping the rest of the prompt fixed, so that anchor and lexical distractor pairs differ in a small, highly local way. 

On the more challenging SCPP benchmark, we observe a different pattern. SCPP introduces systematic edits to objects, attributes, and relations while keeping the rest of the description closely matched, making purely lexical cues much less discriminative. SCPP subtask results are provided in Appendix~\ref{sec:appendix_scpp_results}. It shows that models perform best in cases where disambiguation can be achieved by tracking changes to the main object of the description, but they struggle substantially when only attributes (e.g., color, size) or relations (e.g., subject–object roles, spatial relations) are modified. This highlights an important limitation of current embedding models, which can perform well on easier entity-level contrasts yet still fail on fine-grained semantic changes under high lexical overlap.


Taken together, these results suggest that metric learning approaches tend to exploit the easiest discriminative signal available, often overt differences in entities or salient attributes. This aligns with evidence that contrastive/metric objectives can admit “shortcut” solutions and may not reliably force the model to encode the intended features~\citep{RobinsonSYBJS21}. 



\section{Probing Lexical and Semantic Structure}
\label{sec:lex_semantic_probes}
In this section, we analyze whether the failure patterns of lexical and semantic signals correlate with the 
accessibility of this information in the representations. 
We hypothesize that information that is easy to access should be decodable by a linear classifier. To test this, we train linear probing models to measure the 
lexical and semantic decodability. 
For \textbf{lexical decodability}, the
classifier predicts the corresponding input token from its hidden state (an operational proxy for surface-form information). For \textbf{semantic structure}, we extract sentence embeddings from each layer and evaluate them on MTEB using the benchmark’s standard lightweight heads. As the backbone is kept frozen, differences across layers reflect representational changes rather than task-specific adaptation. Training details for the probes and MTEB dataset are provided in Appendix~\ref{sec:appendix_lex_sem_probe}.


\subsection{Lexical Probes}
\label{sec:lex_probe}

\begin{figure}[!t]
    \centering
    \begin{subfigure}{0.47\textwidth}
        \centering
        \includegraphics[width=\linewidth]{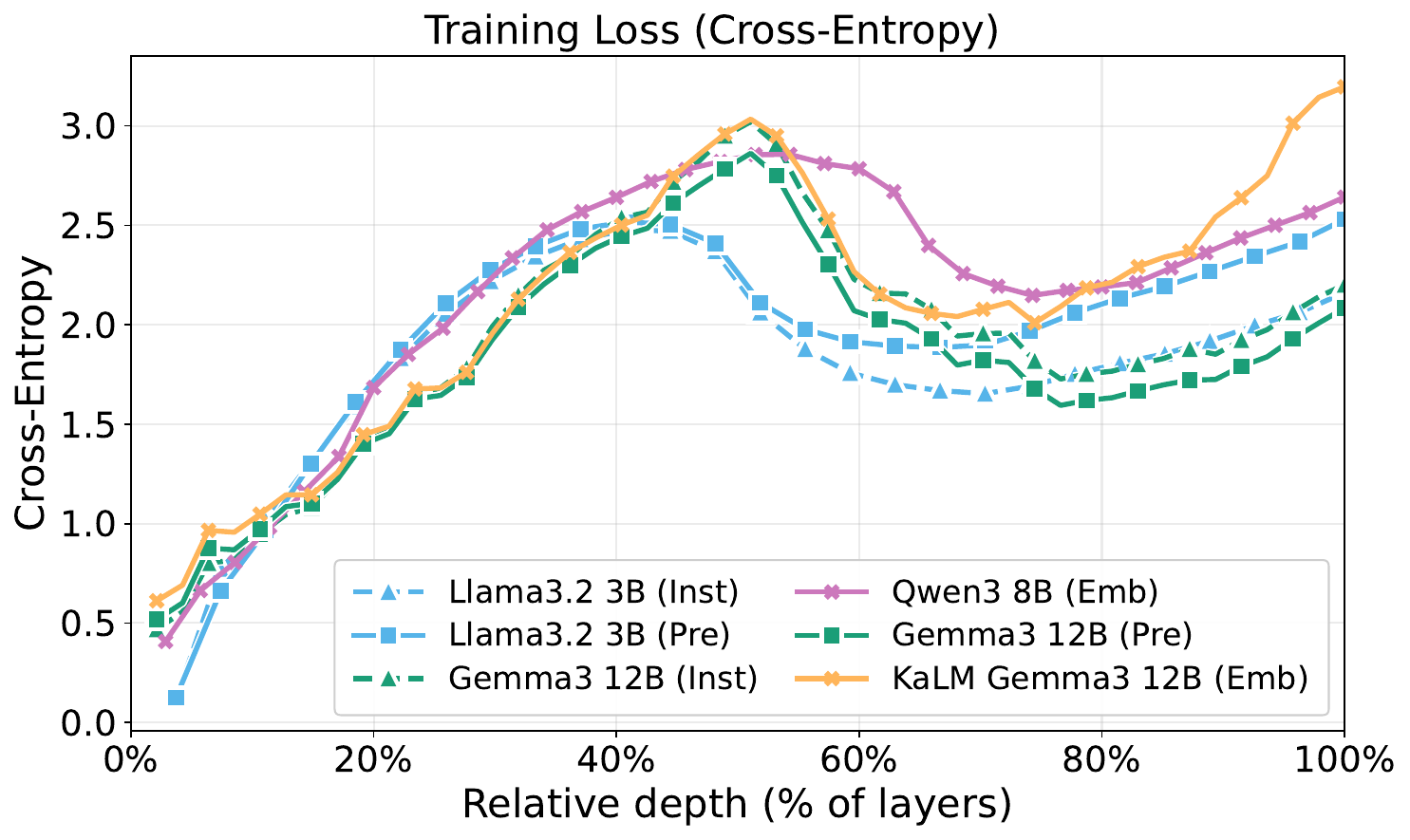}
    \end{subfigure}\hfill
    \begin{subfigure}{0.47\textwidth}
        \centering
        \includegraphics[width=\linewidth]{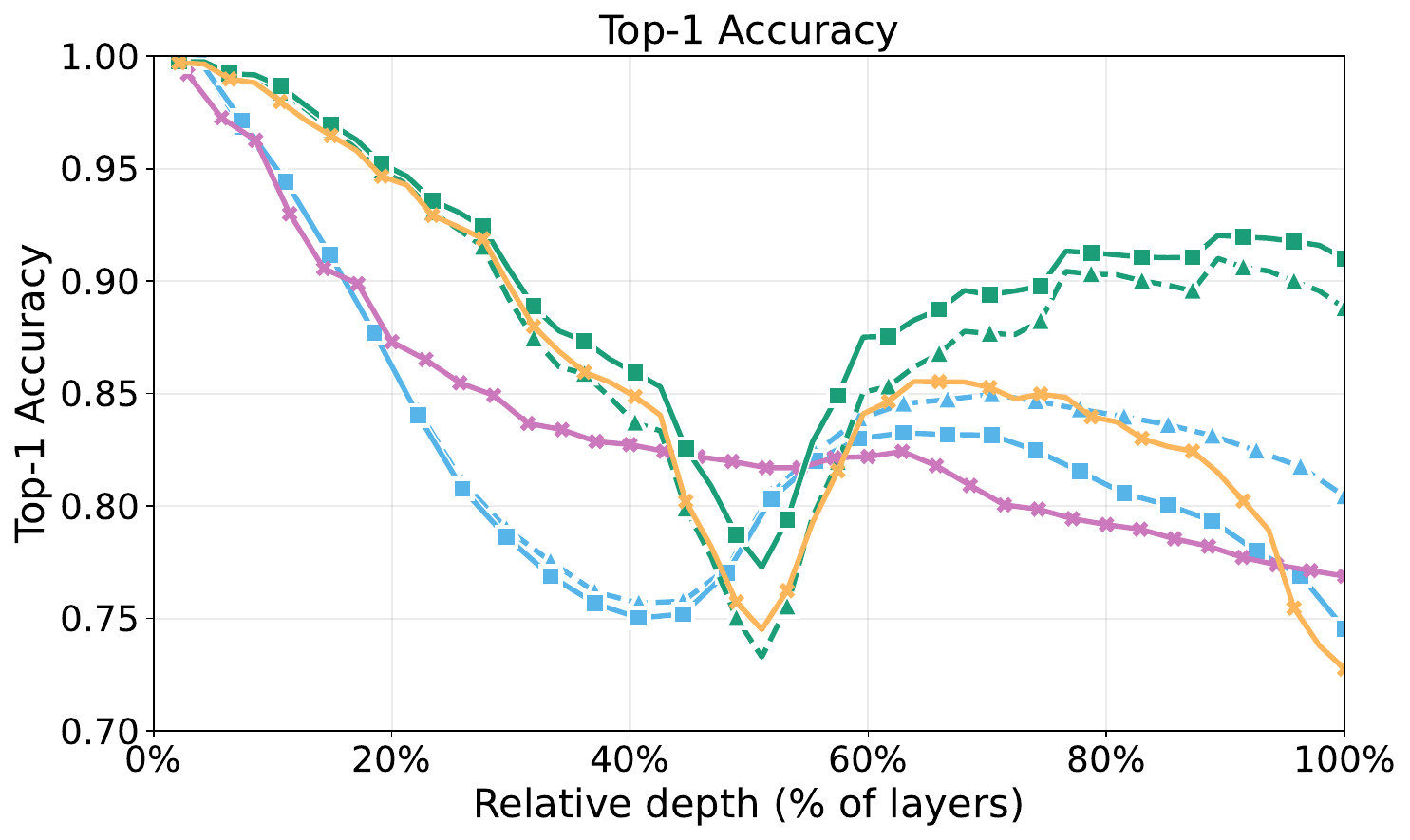}
    \end{subfigure}
    \caption{Layer-wise probe performance for lexical decodability on WikiText across model depths and architectures, measured by cross-entropy loss and Top-1 accuracy.}
    \label{fig:lex_probes}
    \vspace{-10pt}
\end{figure}


For each layer $l \in \mathcal{L}$ and token position $t \in \{1,\dots,T_x\}$, 
$H^{M}_{l}(x)_t \in \mathbb{R}^d$ denotes the model's $d$-dimensional hidden 
representation at position $t$ in layer $l$. To test whether token identity is 
linearly recoverable from this representation, we apply a layer-specific affine 
probe:
\[
    z_{l,t} = W_l H^{M}_{l}(x)_t + b_l,
\]
where $W_l \in \mathbb{R}^{d \times d}$ and $b_l \in \mathbb{R}^d$.
We score each vocabulary token using the model's input embedding matrix 
$E_{\text{in}} \in \mathbb{R}^{V \times d}$, whose $v$-th row 
$e_v \in \mathbb{R}^d$ is the embedding for token $v$. We treat 
$E_{\text{in}}$ as a natural reference basis for surface lexical form, since it 
is the learned lookup table that maps discrete token identities into continuous 
vectors at the input. After $\ell_2$-normalizing both $z_{l,t}$ and the embedding 
rows $e_v$, the score assigned to token $v$ is $
s_{l,t,v} = \langle z_{l,t}, e_v \rangle .
$
Applying a softmax over $v \in \{1,\dots,V\}$ gives a probe-induced distribution 
over token identities. The details regarding training and hyperparameters for the probes is provided in Appendix~\ref{sec:appendix_probe_train_hyperparams}.


Figure~\ref{fig:lex_probes} reports top-1 accuracy and cross-entropy (CE) for token-identity probes as a function of relative depth. 
Token identity is highly linearly decodable in the earliest layers (high accuracy / low CE). As depth increases, probe performance degrades, and cross-entropy rises, peaking around mid-depth, roughly the middle half of the network, indicating that token identity is least linearly accessible in these intermediate representations. Notably, this mid-depth regime coincides with the layers where we observe the highest lexical failure rates on CounterFact and SCPP. This points to a transient re-encoding regime in which sensitivity to lexical influence may persist even as linear decodability of token identity decreases. Prior work links intermediate layers to more syntactic abstractions~\citep{JawaharSS19}, providing a complementary view of this region of the network. Beyond this point, token-identity decodability partially recovers in later layers before deteriorating again near the final layers, where representations are increasingly shaped by the model's training objectives. This non-monotonic pattern cautions against treating depth as a smooth lexical to semantic progression. 



\begin{figure*}[!t]
    \centering
    \begin{subfigure}{0.48\textwidth}
        \centering
        \includegraphics[width=\linewidth]{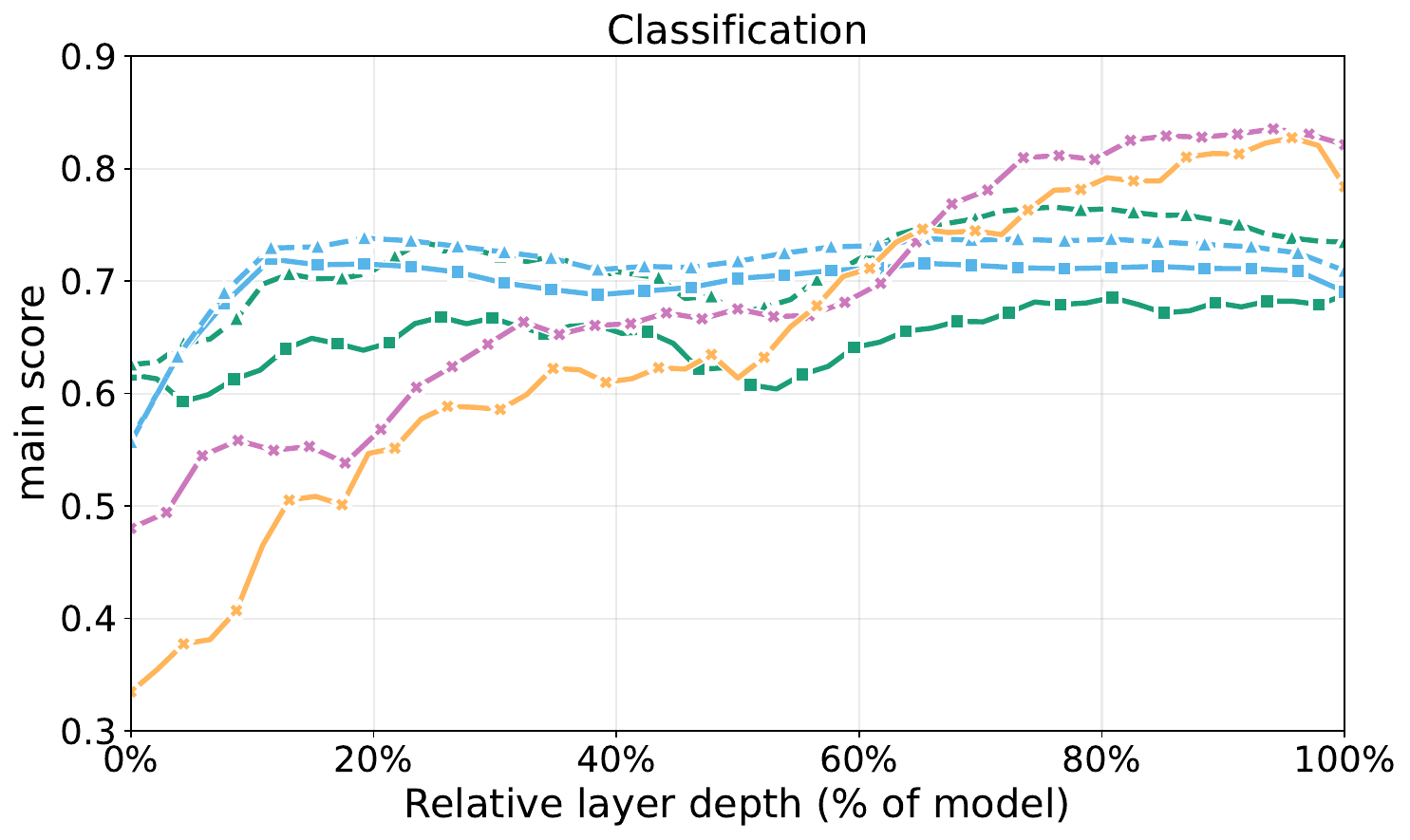}
        \caption{}
    
    \end{subfigure}
   \hfill
    \begin{subfigure}{0.48\textwidth}
        \centering
            \includegraphics[width=\linewidth]{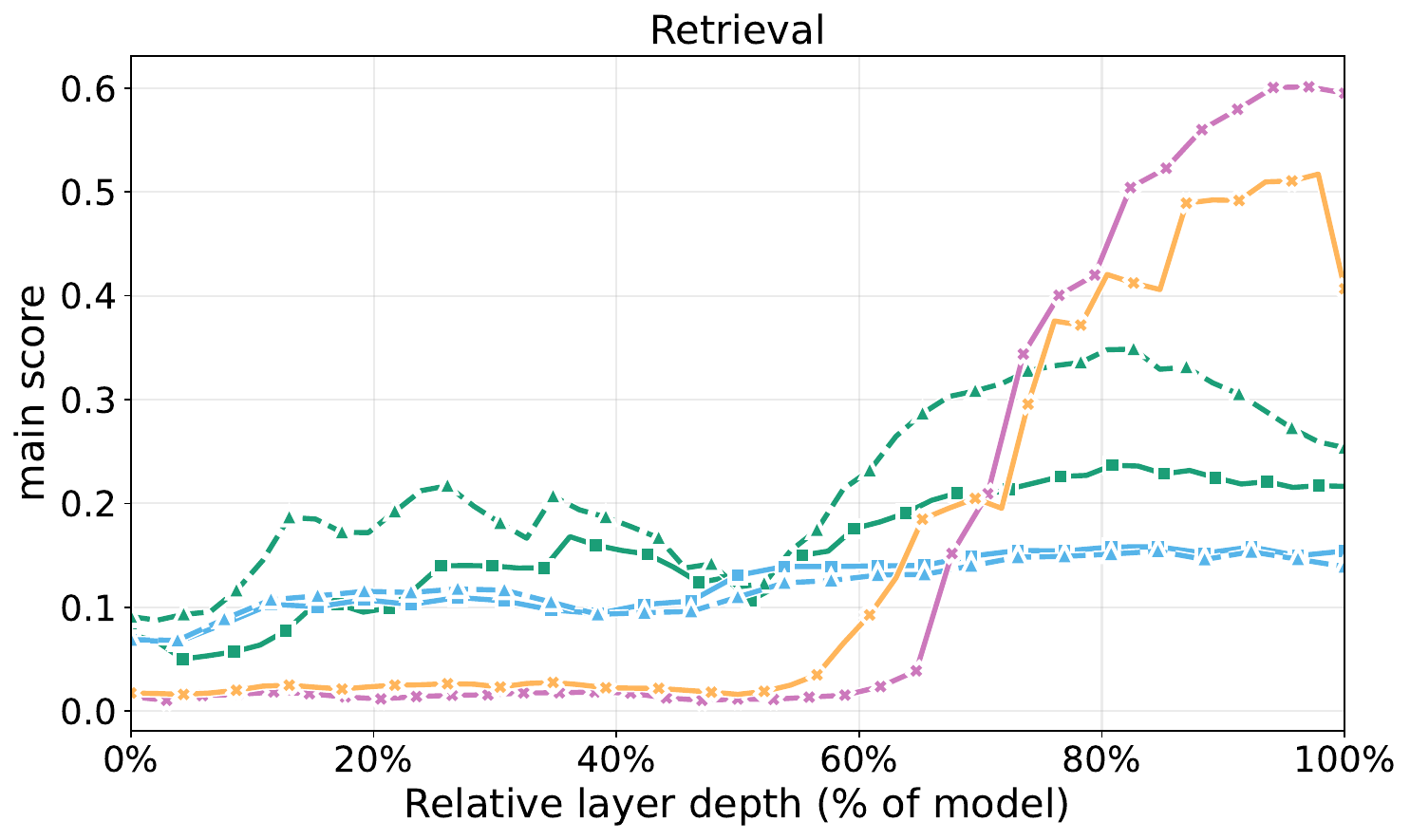}
        \caption{}
    
    \end{subfigure}
    \hfill
    \begin{subfigure}{0.48\textwidth}
        \centering
        \includegraphics[width=\linewidth]{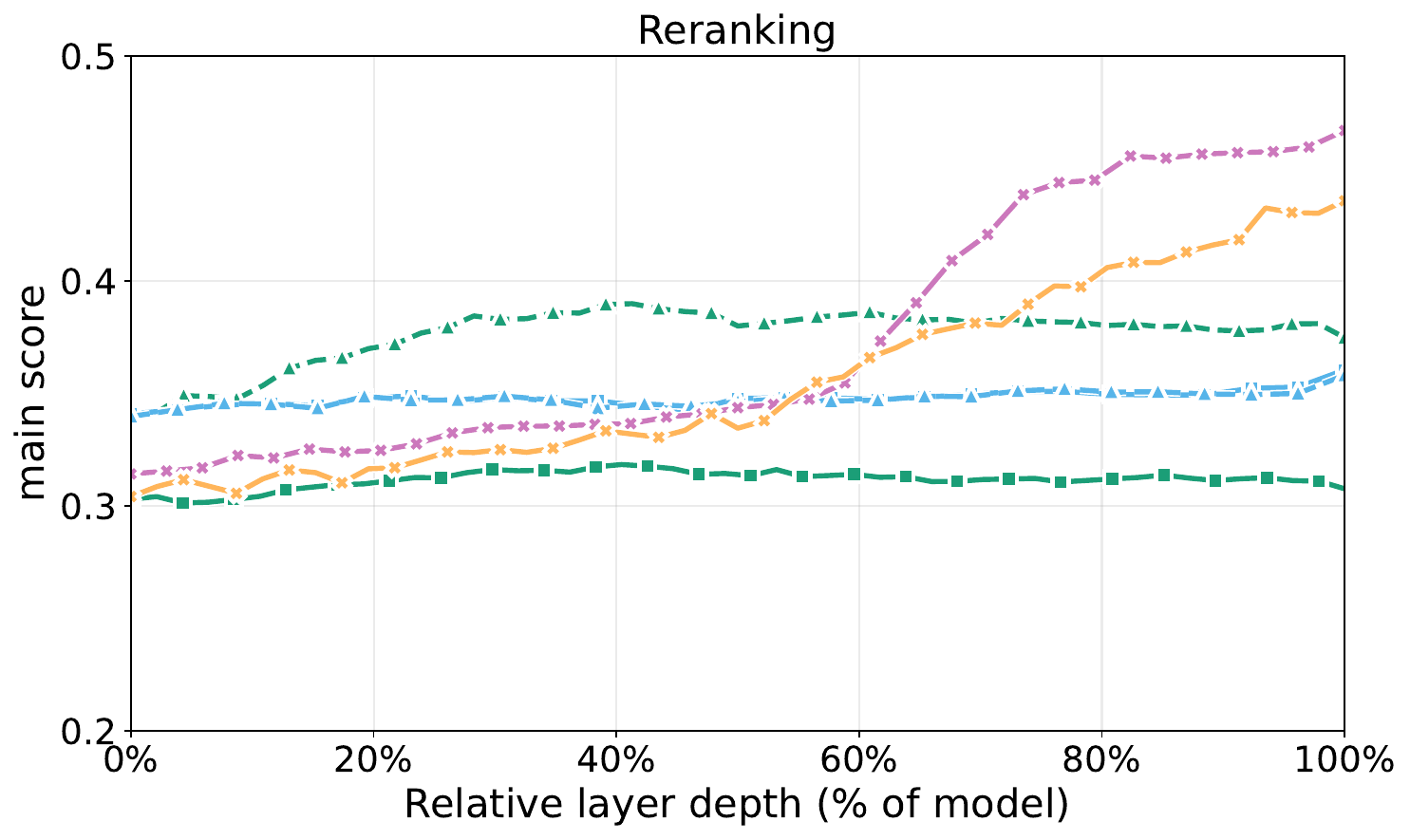}
        \caption{}
    \end{subfigure}
    \hfill
    \begin{subfigure}{0.48\textwidth}
        \centering
        \includegraphics[width=\linewidth]{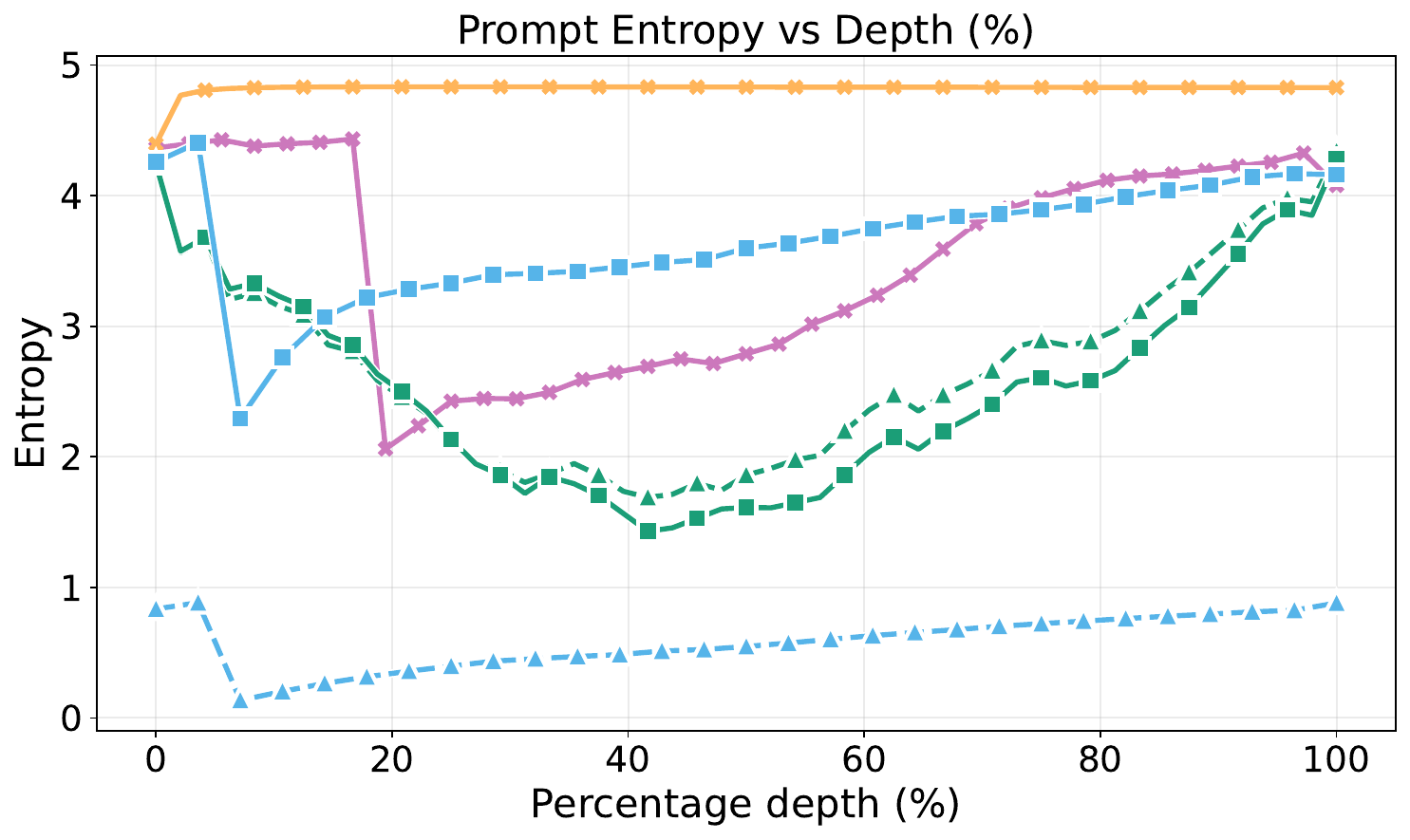}
        \caption{}
    \end{subfigure}
    \begin{subfigure}{\textwidth}
        \centering
        \includegraphics[width=0.98\textwidth]{mteb_legend.pdf}
    \end{subfigure}
    \caption{MTEB task performance across the full model depth for all evaluated models. For each dataset category, curves show the average task performance as a function of depth $l$, using layer-wise sentence embeddings.}
    \label{fig:semantic_probes_entropy}
    \vspace{-10pt}
\end{figure*}

\subsection{Semantic Probes
}







We follow MTEB’s two complementary evaluation settings to probe semantic information in layerwise sentence embeddings. MTEB measures performance either directly from embedding similarities and induced rankings (no task-specific training) or by training a lightweight linear probe to test whether task-relevant information is linearly separable at a given layer. Dataset and evaluation details are provided in Appendix~\ref{sec:appendix_semantic_probe}.

Figure~\ref{fig:semantic_probes_entropy} reports MTEB results for classification (a), retrieval (b), and reranking (c). We provide the results for clustering, pairwise classification, and STS in Appendix~\ref{sec:appendix_mteb_results}. 
Although MTEB spans diverse tasks that could yield heterogeneous layerwise trends, we see strong regularity where performance varies with depth, and curve trends cluster into a few characteristic profile types.
Across classification, pairwise classification, retrieval, and semantic textual similarity (STS), we consistently observe a non-monotonic, valley-shaped depth profile in all models:
performance improves from early to mid layers, degrades in an intermediate regime, and then recovers near the top of the model. In contrast, clustering and reranking do not exhibit the mid-depth valley observed for classification-style tasks. For reranking, performance remains essentially constant across depth. Clustering generally improves with depth, but the trend is noisy under the default MTEB setup (k-means + V-measure), where discrete assignment changes can cause jagged layer-to-layer variation.

For \textit{embedding models}, performance in the intermediate ``valley'' regime either remains roughly constant or improves slowly, before increasing sharply after this. Peak scores for all models typically occur in the last few layers, but not consistently at the final layer, suggesting that the standard “use the final layer” heuristic can be suboptimal for semantic benchmarks, consistent with findings by~\citet{SkeanAZPNLS25,chengemergence}. 

Together with \S\ref{sec:measure_lex_infl}, these results identify a shared mid-depth transition: lexical features become least recoverable over a broad intermediate regime, without a compensating semantic gain in MTEB. Performance is often lowest over the same layers and recovers only later, suggesting a phase where surface form is suppressed before semantic structure is fully organized. Consequently, the strongest representations typically emerge in mid-to-late layers.




\section{An Information-Theoretic Lens on the Mid-Depth Valley}
\label{sec:input_entropy}

To further explore how lexical and semantic structure evolve across depth, we juxtapose our probing results with an information-theoretic view of representations, which frames learning as a trade-off between compression and the preservation of predictive structure~\citep{tishby2000information,shwartz2017opening}. Geometric accounts based on intrinsic dimension reach a compatible conclusion from a different direction: hidden representations do not evolve monotonically with depth but instead pass through distinct regimes of expansion, contraction, and abstraction~\citep{valeriani2023geometry,chengemergence,DoimoSAC24}. \citet{SkeanAZPNLS25} offer a particularly relevant comparison, showing that Transformer representations follow a characteristic compression–decompression trajectory across layers and arguing that the low-entropy intermediate regime hosts the strongest representations.


To connect our probing results to representation geometry, we adopt prompt entropy as a per-layer summary statistic (formal definition in Appendix~\ref{sec:appendix_entropy}). Intuitively, low entropy indicates compressed, low-effective-rank token representations.
Figure~\ref{fig:semantic_probes_entropy}(d) confirms the expected shape: entropy falls into the middle layers and rises again toward the output.

Analysing the entropy curves and empirical results in \S~\ref{sec:measure_lex_infl} and \S~\ref{sec:lex_semantic_probes} together provides a more precise account of where the strongest representations emerge. The mid-depth valley at 40--60\% relative depth aligns with the low-entropy bottleneck and its immediate re-expansion phase, precisely the regime that \citet{SkeanAZPNLS25} associates with the strongest intermediate representations. Yet in our measurements, this is where semantic performance is weak and lexical influence is most pronounced: the model is maximally compressed but not maximally meaningful. Clean semantic separation and reduced lexical-overlap failures only emerge in the later layers, after entropy has begun to rise. This nuances the compression-centric account: the low-entropy bottleneck is not the point of strongest embedding quality. Instead, semantic performance is weak, and lexical influence is most pronounced near the bottleneck and the early re-expansion phase. These findings suggest that the re-expansion phase plays a critical role in producing a more meaning-sensitive embedding geometry.

\section{Practical Implications}
\label{sec:practical_implications}
In this section, we explore the impact of lexical influence on practical settings that rely on embedding geometry and representation-sensitive updates.

\subsection{Summarization}


Text summarization compresses a document while preserving its key content. The evaluation of summarization involves the use of both n-gram-based metrics and semantic metrics. The former, such as ROUGE~\citep{lin2004rouge} and BLEU~\citep{papineni2002bleu}, reward n-gram overlap as a signal for a better summary, emphasizing phrase matching in contrast to semantic matching. The latter metrics, such as BERTScore and BARTScore, aim at evaluating summaries based on meaning preservation. Here, we show that despite the focus of evaluation methods on semantics, the underlying models possess lexical bias that impacts the correct semantic evaluation of summaries.

\textbf{Data.} We use the 
Extreme Summarization benchmark~\cite{narayan2018don}. To probe lexical sensitivity, we augment each example with a highly lexicalized non-summary distractor generated by an LLM. Details for dataset construction and experimental methodology are provided in Appendix~\ref{sec:summarization_data_evaluation}.

\textbf{Experiment.} For each document, we score (i) the gold summary and (ii) an LLM-generated lexical distractor with \textit{BERTScore} and \textit{BARTScore}. We count a \emph{failure} when the distractor scores at least as high as the gold summary. BERTScore fails on $64\%$ of examples, consistent with a strong lexical signal in contextual representations,  mirroring our earlier findings. BARTScore fails on $20\%$: despite being likelihood-based, lexical distractors can remain locally plausible and score competitively. This shows that lexical reliance is not limited to contextual representation similarity metrics, and the token-level generative process can reward lexically aligned text as well.

\begin{table*}[t]
\centering

\begin{minipage}[t]{0.47\textwidth}
\centering
\footnotesize
\caption{Summarization ranking results (lexical distractor vs.\ gold). FR is the fraction of cases where the distractor is scored higher than the gold; ASD/ASG are mean scores for distractor/gold (method-specific scales).}
\setlength{\tabcolsep}{6pt}
\renewcommand{\arraystretch}{1.15}
\begin{tabular*}{\linewidth}{@{\extracolsep{\fill}} l c c c @{}}
\toprule
\textbf{Method} & \textbf{FR} $\downarrow$ & \textbf{ASD} $\downarrow$ & \textbf{ASG} $\uparrow$ \\
\midrule
BERTScore & 0.640 & 0.691  & 0.649 \\
BARTScore & 0.200 & -3.918 & -2.945 \\
\bottomrule
\end{tabular*}
\label{tab:method_results}
\end{minipage}
\hfill
\begin{minipage}[t]{0.47\textwidth}
\centering
\footnotesize
\caption{Mean (over prompts) distribution-shift metrics for lexically similar (LS) and lexically dissimilar (LD) prompts.}
\setlength{\tabcolsep}{4pt}
\renewcommand{\arraystretch}{1.12}
\begin{tabular*}{\linewidth}{@{\extracolsep{\fill}} l c c c c @{}}
\toprule
\multirow{2}{*}{\textbf{Metric}} &
\multicolumn{2}{c}{\textbf{Top-K-20}} &
\multicolumn{2}{c}{\textbf{Top-K-50}} \\
\cmidrule(lr){2-3}\cmidrule(lr){4-5}
& \textbf{LS} & \textbf{LD} & \textbf{LS} & \textbf{LD} \\
\midrule
JSD            & 0.290 & 0.242 & 0.293 & 0.245 \\
TV             & 0.561 & 0.493 & 0.565 & 0.499 \\
Kendall $\tau$ & -0.016 & 0.030 & -0.022 & 0.028 \\
\begin{tabular}[c]{@{}l@{}}Edited-Target\\Presence\end{tabular}
               & 0.023 & 0.021 & 0.085 & 0.043 \\
\midrule
\multicolumn{3}{l}{Top-1 Flip Rate \textit{(Top-K-indep.)}} & 0.566 & 0.493 \\
\bottomrule
\end{tabular*}
\label{tab:editing_results}
\end{minipage}

\vspace{-9pt}
\end{table*}

\subsection{Model Editing}
Deploying LLMs in production requires post-training updates to maintain correctness and policy compliance (e.g., incorporating new facts or suppressing harmful associations)~\cite{MengBAB22,mitchell2021fast}. However, retraining is costly~\citep{hartvigsen2023aging,RizwanRW025}, and fine-tuning can induce catastrophic forgetting ~\citep{luo2025empirical,wang2024knowledge}, while prompt-based methods require routing and are brittle~\cite{rosati2024long,su2025parametric}. Model editing addresses this gap by enabling targeted, localized updates that satisfy an edit request while preserving behavior on non-target inputs.

Formally, given a model $M_{\theta_0}$, an edit request $(x_e, y_e^\star)$, and a locality set $\mathcal{N}$ representing all other behaviors that should remain unchanged, model editing seeks either updated parameters $\theta'$ (weight-modifying) or an auxiliary component parameterized by $\phi$ (e.g., a learned prefix, adapter, or external memory) while keeping the base weights $\theta_0$ fixed (weight-preserving), such that the edit succeeds while non-target behavior is preserved:
\[
\begin{aligned}
& M_{\theta'}(x_e)=y_e^\star \;\;\text{or}\;\; M_{\theta_0,\phi}(x_e)=y_e^\star, \\
&  D\!\left(\bigl(M_{\theta'} \,\big|\, M_{\theta_0,\phi}\bigr)(\cdot \mid x),\,M_{\theta_0}(\cdot \mid x)\right)\le \varepsilon
\quad \forall x\in\mathcal{N}.
\end{aligned}
\]
\citet{RizwanRW025} showed that semantic-similarity based scoping systems can fail in weight-preserving editing due to lexical overlap. We argue that this failure extends to \emph{weight-modifying} editors: surface-token similarity can bring semantically unrelated prompts close in embedding space, so an update targeted at $x_e$ also shifts their output distributions. Consequently, lexically similar but semantically mismatched distractors are disproportionately perturbed, producing systematic lexical skew in locality.


\textbf{Method and dataset.}
We use AlphaEdit~\citep{FangJWMSW0C25} and evaluate on a modified version of the CounterFact benchmark. For each edit, we augment CounterFact with two prompt sets: (\emph{i}) \emph{lexically similar} variants created by substituting distractor entities into the edit prompt, and (\emph{ii}) \emph{lexically dissimilar} variants that keep the distractor's semantics but substantially change its surface form.

\textbf{Metrics.} 
We quantify distribution shift at the answer position by comparing the base and edited next-token distributions. We compute Jensen–Shannon divergence (JSD) and total variation (TV), i.e.
probability change, Kendall's $\tau$ for rank stability, and the Top-1 Flip Rate; all metrics are evaluated on the merged top-$K$ support. The details for method, dataset, metrics, and $\text{top-K}=100$ results are provided in Appendix~\ref{sec:appendix_model_editing}. To measure leakage into distractors, we report \textsc{Edited-Target Presence}: the fraction of locality prompts whose edited-model top-$K$ contains any relation-specific edited target token.


Table~\ref{tab:editing_results} reports the results of our locality experiment. Across metrics and Top-$K$ settings, lexically similar (LS) locality prompts are consistently more affected by the edit than lexically dissimilar (LD) locality prompts, even though both prompt families query facts unrelated to the edited target. The effect is stable across $K$, indicating it is not an artifact of the Top-$K$ cutoff.

The effect is clearest in the probability-based shift metrics. Relative to LD, LS exhibits larger changes in both the shape of the distribution (JSD) and the amount of probability mass moved (TV), implying that surface-form overlap amplifies the edit’s collateral redistribution over plausible candidates. Rank-based evidence is consistent with this picture: Kendall’s $\tau$ shows greater rank instability under LS, with the separation most apparent at smaller $K$, where $\tau$ more directly reflects reordering among high-likelihood tokens. As $K$ increases, $\tau$ becomes less discriminative because it averages over many additional token pairs, which can dilute head reordering by including comparatively stable mid and lower-probability tokens. Finally, behavioral indicators align with these distributional shifts: LS distractors exhibit more decision-level changes (Top-1 flips) and more frequent appearance of relation-specific edited targets in the post-edit top-$K$, consistent with stronger edit leakage under lexical overlap.

\textbf{Implication.} The results indicate that surface-form overlap systematically increases a prompt’s susceptibility to unintended changes. More broadly, this suggests that any evaluation criterion for model interventions (e.g., steering or activation-based methods, adapters/LoRA, targeted fine-tuning) should stress-test scope using lexically similar control prompts that request different information.



\section{Related Work}


Layer-wise probing studies show that linguistic properties are not uniformly encoded across depth, but peak at different layers \citep{Liu0BPS19,TenneyDP19,JawaharSS19}; later work strengthens these analyses with selectivity and control tasks \citep{HewittL19}. Complementary geometric and similarity-based approaches study how embedding spaces themselves evolve across layers, including changes in contextual geometry and representational alignment \citep{Ethayarajh19,Kornblith0LH19}.
\citet{valeriani2023geometry} further characterizes the geometry of hidden representations in large transformers, showing that intrinsic dimension and neighborhood structure vary systematically across depth. Recent work extends this with unified depth diagnostics and large-scale evaluations showing that intermediate layers can be especially informative feature sources~\citep{SkeanAZPNLS25}, as well as evidence for high-dimensional abstraction phases~\citep{chengemergence}, staged inference dynamics~\citep{lad2406remarkable}, and representation-landscape changes induced by few-shot learning and fine-tuning \citep{DoimoSAC24}. Related depth-centric work further shows that concepts and knowledge emerge progressively across layers \citep{haider2024looking, JinYHZWH0MMDYDZ25}, while predictions can stabilize well before the final layers \citep{SajjadDDN23, fan2025}.


Lexical bias is often framed as shortcut learning, where models exploit surface cues that correlate with labels rather than 
semantics. In NLI, annotation artifacts introduce exploitable lexical patterns
\citep{GururanganSLSBS18,HeZW19,PoliakNHRD18}. 
Controlled diagnostics like  HANS~\citep{McCoyPL19} show that high performing 
models often follow lexical overlap and subsequence heuristics and fail when these heuristics are broken. Large scale 
evaluations further confirm 
vulnerabilities to lexical shortcuts 
\citep{YuanZZZL24}. Lexical shortcuts can persist even after mitigation, motivating debiasing objectives that target unknown biases rather than a manually specified cue \citep{SerranoDS23,UtamaMG20,zhou2020towards}. 

\section{Conclusion}
We study the persistent effects of lexicality on LLM representations by tracking how lexical and semantic structure evolve across depth. We find that instruction tuning and metric-learning objectives improve semantic performance, but do not necessarily overcome lexical influence, so lexical-overlap failures can persist despite gains on standard embedding evaluations: even embedding models trained for semantic similarity struggle on fine-grained attribute and relation changes under high lexical overlap. Connecting the stress test with layer-wise token-identity probes and semantic evaluations across model families and training regimes, we uncover a consistent mid-layer valley in which semantic fidelity remains nearly constant while overlap-driven failures peak, even as token identity is least linearly decodable. Reading these results alongside the entropy curve refines the compression-centric view, showing that the low-entropy bottleneck is not where the strongest semantic representations arise. Instead, stronger meaning-sensitive geometry emerges later, during re-expansion. We further show that this lexical influence propagates into downstream LLM usage, as demonstrated with the use cases of automatic summarization evaluation and model editing. 

We conclude that lexically controlled stress tests should become a standard check for representation learning, automatic metrics, and model interventions
because they reveal whether meaning sensitivity is robust to high-overlap confounds or breaks down through overlap-driven semantic failures.
In future work, we will dissect the mid-depth transition regime with nonlinear transformations to identify what structure is formed there and how to prevent overlap-driven similarity errors.

\section*{Acknowledgment}
We acknowledge the support of the Natural Sciences and Engineering Research Council of Canada (NSERC), Canada Foundation of Innovation (CFI), and Research Nova Scotia. Advanced computing resources are provided by ACENET, the regional partner in Atlantic Canada, and the Digital Research Alliance of Canada.

\bibliographystyle{plainnat}
\bibliography{neurips_2026}

\appendix

\section{Limitations}
\label{sec:limitations}
Our internal analysis probes representations only at transformer block outputs, so we do not distinguish contributions from finer-grained components (e.g., attention vs. MLP pathways) or alternative hook points. Finally, the probing results are diagnostic: they indicate what information is recoverable from representations, but do not by themselves establish a causal mechanism.

\section{Impact Statement}
\label{sec:impact_statement}
This work characterizes lexical influence in representation spaces, showing that surface-form overlap can distort embedding similarity and miscalibrate similarity-based evaluation under lexical confounds. It also introduces diagnostics that quantify when and where semantic structure degrades. We expect the primary impact to be positive: improving evaluation rigor and informing more robust use of embeddings in retrieval, ranking, and monitoring pipelines. While the findings could in principle be used to stress similarity-based systems via lexically similar distractors, such misuse is likely limited in practice because it generally requires model and pipeline-specific knowledge and sustained control over inputs; more often, the results will help identify and mitigate unintended brittleness.

\section{LLM Usage}
\label{sec:llm_use}
LLMs were used only to improve grammar, wording, and prose clarity. All technical content, experimental design, analysis, and conclusions were developed and verified by the authors.

\section{Datasets}
\label{sec:datasets}

For all experiments in the paper, we prioritize large evaluation sets and aggregate layer-wise trends over repeated split-based reruns, given the breadth of the study across models, tasks, and layers.
\subsection{Stress Tests and Model Editing}
\label{sec:appendix_stress_editing_data}
Table~\ref{tab:datasets_lexical} shows representative examples from CounterFact and SugarCrepe++ (SCPP). Each instance specifies a factual association of the form (subject, relation, object)
 together with a set of prompts designed to elicit that fact. It includes (i) \emph{anchor} prompts that directly query the target relation for the subject, (ii) \emph{paraphrase} prompts that ask for the same fact using alternative surface forms to test generalization beyond a single template, and (iii) \emph{distractor} prompts that query the same relation for semantically related subjects. For model editing, distractors are used to test specificity/locality (i.e., whether the edit remains localized). After an edit, a successful model should produce the new target object for the anchor/paraphrase prompts while leaving predictions on distractor prompts largely unchanged.

SCPP provides image–caption triplets with two semantically equivalent but lexically different positives and a hard negative that is lexically confusable yet semantically incorrect for the image. We utilize only the texts for our paper. It partitions negatives into five subsets based on the perturbation used to create the distractor: \emph{Swap Object} exchanges two object nouns (or noun phrases) in the caption; \emph{Swap Attribute} swaps attributes between objects (e.g., colors/sizes); \emph{Replace Object} substitutes an object noun with another; \emph{Replace Attribute} substitutes an attribute term with another; and \emph{Replace Relation} substitutes the relation predicate (e.g., action or spatial preposition), altering who-does-what-to-whom or the spatial configuration.

\subsection{Stress Test Sample Details}
\label{sec:appendix_sample_details_stress}
For the adversarial stress tests, we use the full SCPP dataset, consisting of 4,752 samples, together with 5,000 samples from CounterFact. CounterFact exists in several variants, including the original version released with ROME~\citep{MengBAB22}, as well as later sampled or updated versions from EasyEdit~\citep{wang2024easyedit} and PENME~\citep{RizwanRW025}. The original dataset contains two paraphrase prompts per sample, whereas the EasyEdit version contains one paraphrase prompt and one locality/neighborhood prompt per sample. The generation prompts contained in the dataset are not exact paraphrases, but rather prompts that may lead to the edited answer. For example, for the edit prompt ``Autonomous University of Madrid, which is located in,'' one generation prompt is ``The best restaurants around Autonomous University of Madrid include.'' Since this is not an exact paraphrase, we cannot use these prompts.

We therefore use the version released by PENME~\citep{RizwanRW025}, which provides the two original paraphrases along with several paraphrases generated via LLM (3 minimum per sample). There are $10$ locality/neighborhood prompts for each sample. For this dataset, we used n-grams to rank locality prompts by lexical similarity to the edit prompt to obtain a clear separation between five lexically similar samples and five dissimilar ones. Using this ranking, we sampled 5,000 samples from the dataset. For the model editing experiments, we used this sampled dataset. For the stress tests, we restricted evaluation to the lexically similar locality prompts, resulting in 25 triplet evaluations per sample and $125,000$ evaluations in total. 

\begin{table*}[!t]
  \centering
  \scriptsize
  \caption{The table shows samples from CounterFact and SugarCrepe++ (SCPP) datasets.}
  \label{tab:datasets}
  \begin{tabularx}{\textwidth}{lXXX}
    \toprule
    \textbf{Dataset} & \textbf{Prompt/Anchor} & \textbf{Paraphrase} & \textbf{Distractor} \\
    \midrule
    CounterFact & Justin Fleming was originally from
    & Justin Fleming was native to & Andrew Durante is originally from\\
    SCPP++ replace attribute & People standing near a table with open pizza boxes & Individuals are positioned close to a table with open pizza boxes. & People standing near a table with closed pizza boxes. \\
    SCPP++ replace relation & A motorcycle is parked inside of a building. & The building contains the parked motorcycle. & A motorcycle is parked outside of a building. \\
    SCPP++ swap attribute & A person opens a fire hydrant and brown colored water is flowing out. & Brown-colored water flows out of the fire hydrant opened by a person. & A person opens a brown colored hydrant and water is flowing out. \\
    SCPP++ replace object & Multiple time pieces displayed on the wall and on a flat surface. & Several timepieces are exhibited on a surface which is flat, and on the wall. & Multiple time pieces displayed on the ceiling and on a flat surface. \\
    SCPP++ swap object & A child holds a spoon and looks at a cupcake. & The child looks at a cupcake while holding a spoon. & A child looks at a spoon and holds a cupcake. \\
    \bottomrule
  \end{tabularx}
  
  \label{tab:datasets_lexical}
\end{table*}

\begin{figure*}[!t]
    \centering
    \begin{subfigure}{0.32\textwidth}
        \centering
        \includegraphics[width=\linewidth]{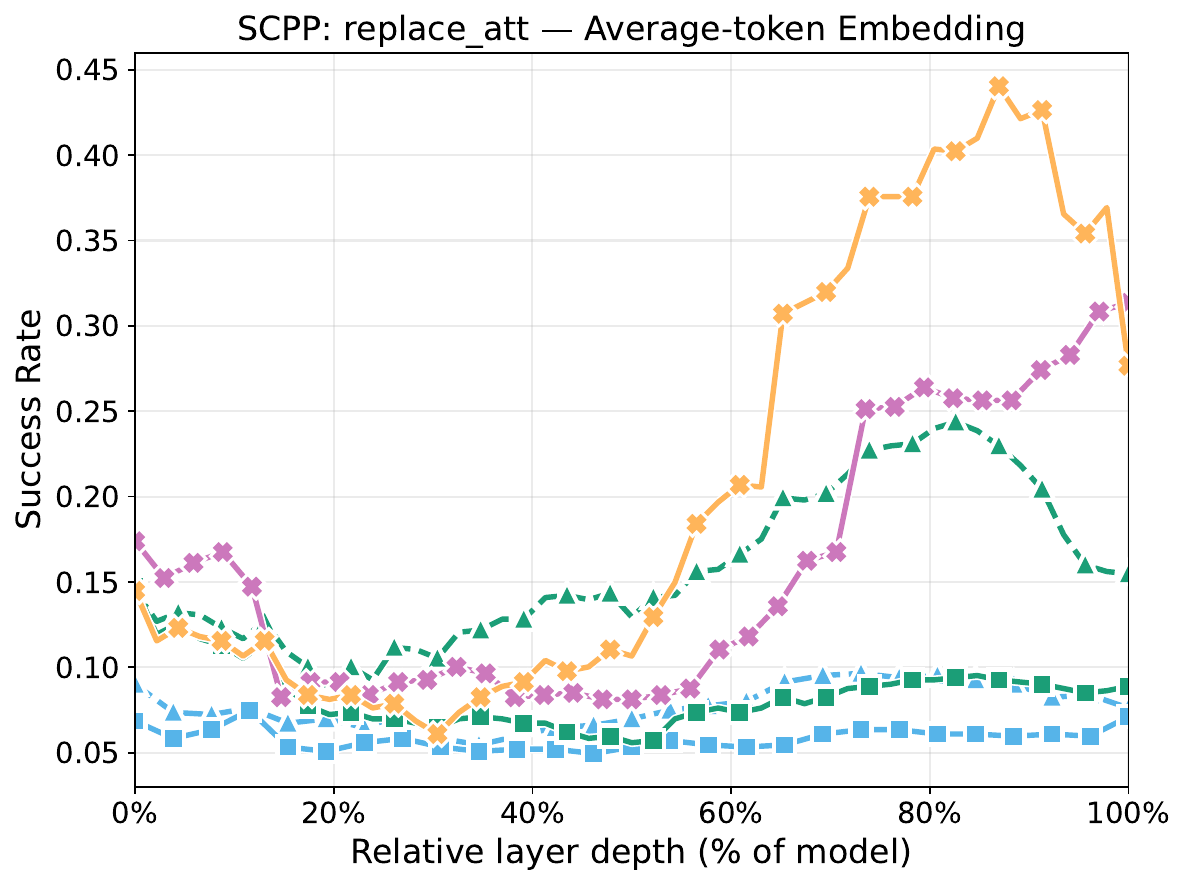}
    \end{subfigure}
    \hfill
    \begin{subfigure}{0.32\textwidth}
        \centering
        \includegraphics[width=\linewidth]{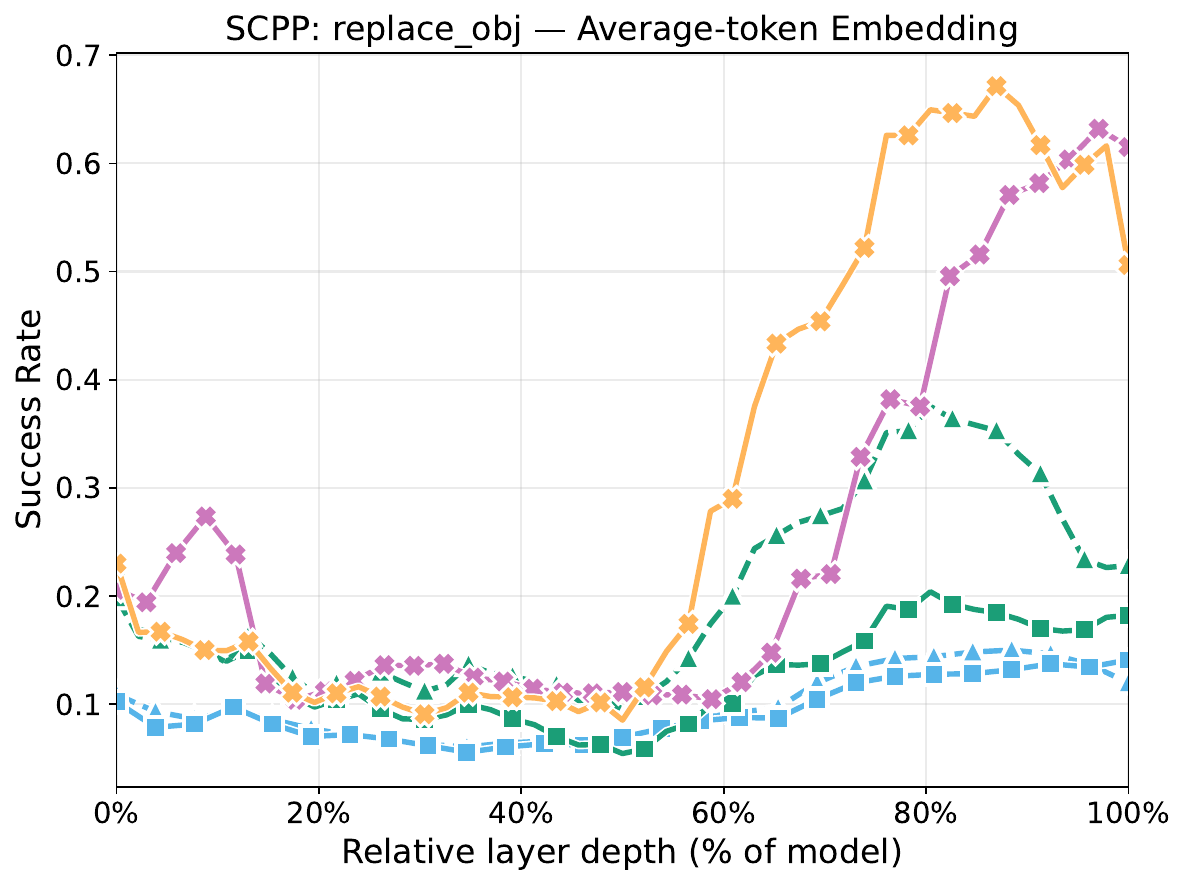}
    \end{subfigure}
    \begin{subfigure}{0.32\textwidth}
        \centering
        \includegraphics[width=\linewidth]{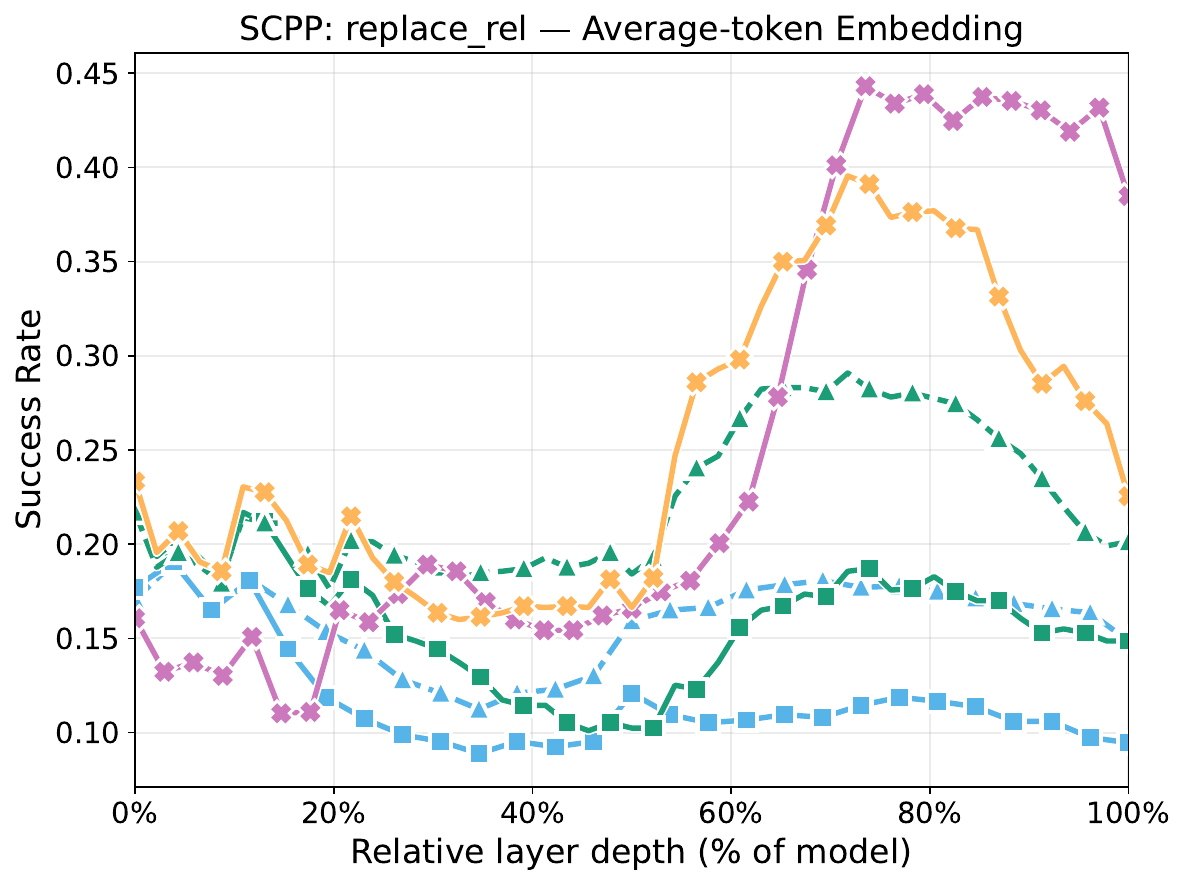}
    \end{subfigure}

    \begin{subfigure}{0.32\textwidth}
        \centering
        \includegraphics[width=\linewidth]{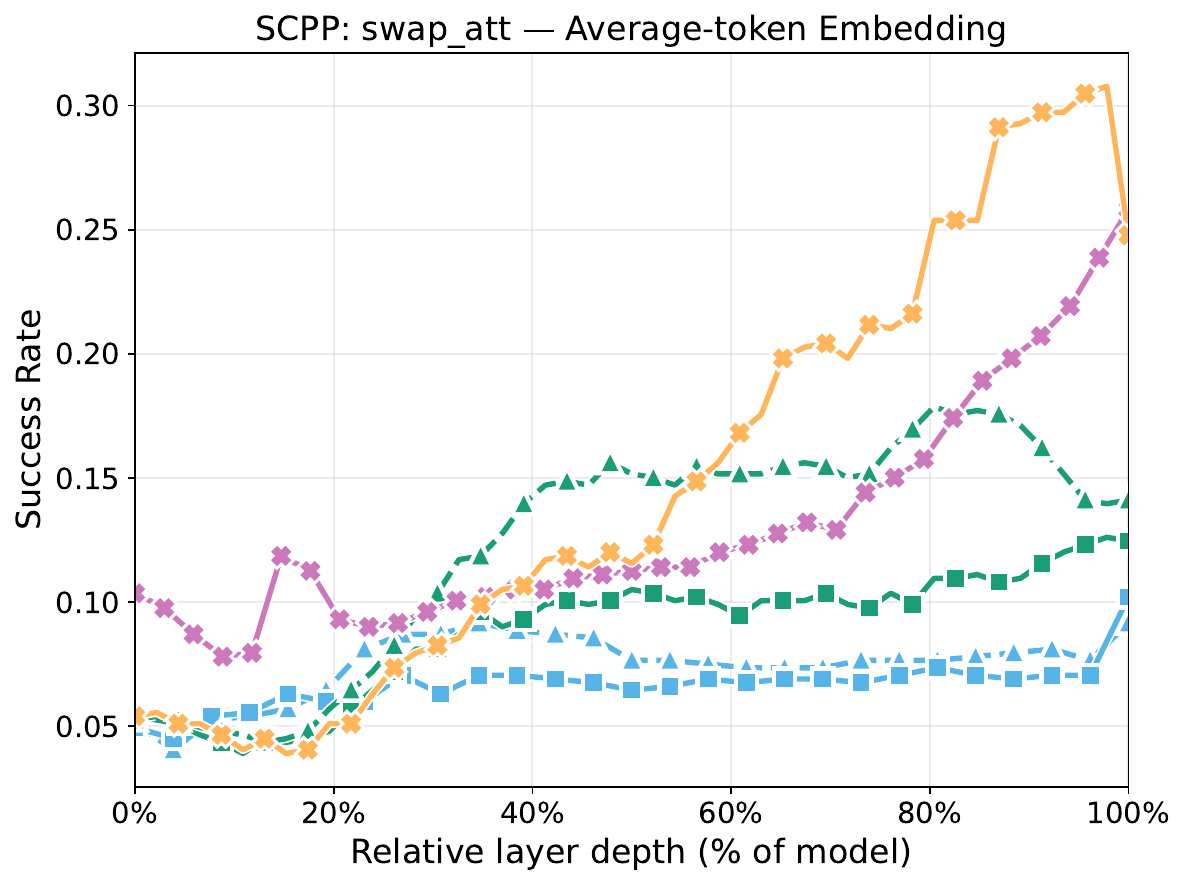}
    \end{subfigure}
    \begin{subfigure}{0.32\textwidth}
        \centering
        \includegraphics[width=\linewidth]{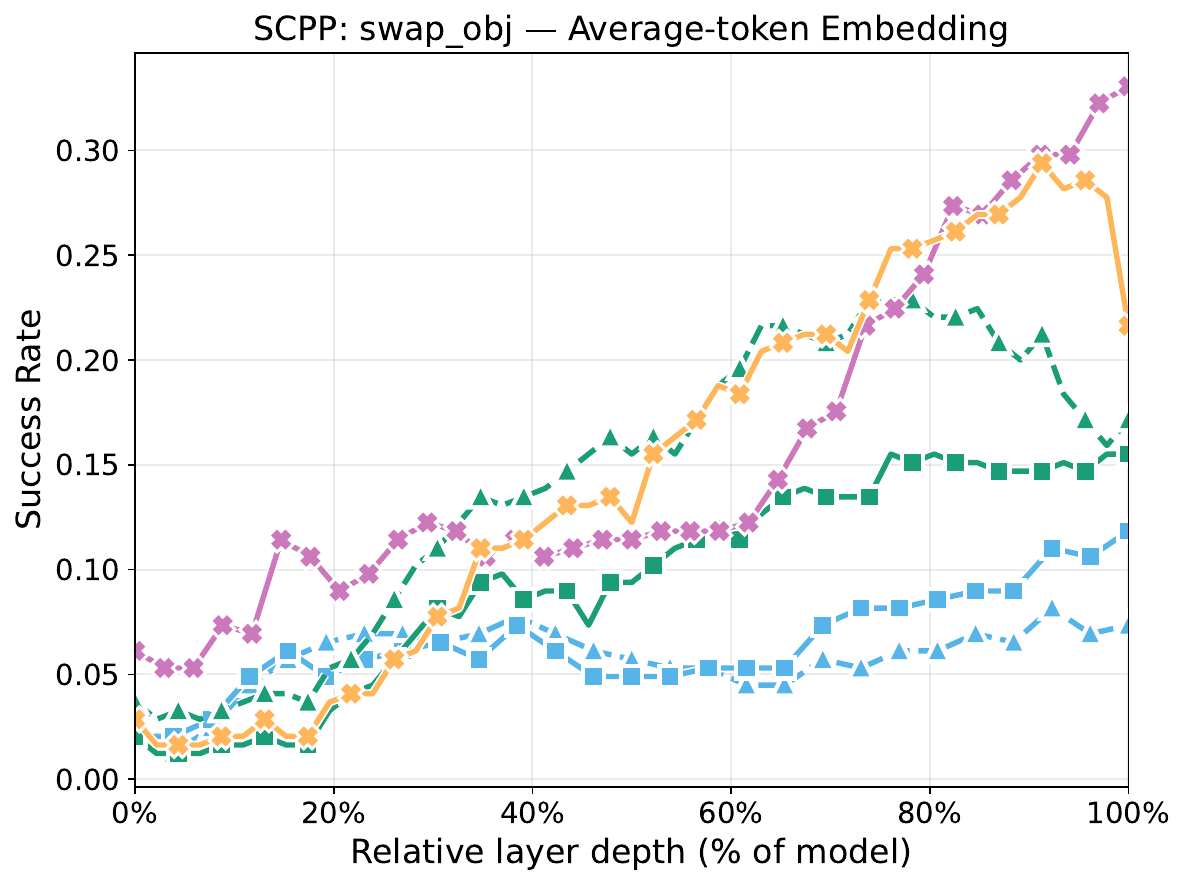}
    \end{subfigure}
    \begin{subfigure}{0.9\textwidth}
        \centering
        \includegraphics[width=\linewidth]{mteb_legend.pdf}
    \end{subfigure}

    \caption{Results for SCPP average token embeddings}
    \label{fig:appendix_scpp_average}
\end{figure*}

\begin{figure*}[!t]
    \centering
    \begin{subfigure}{0.32\textwidth}
        \centering
        \includegraphics[width=\linewidth]{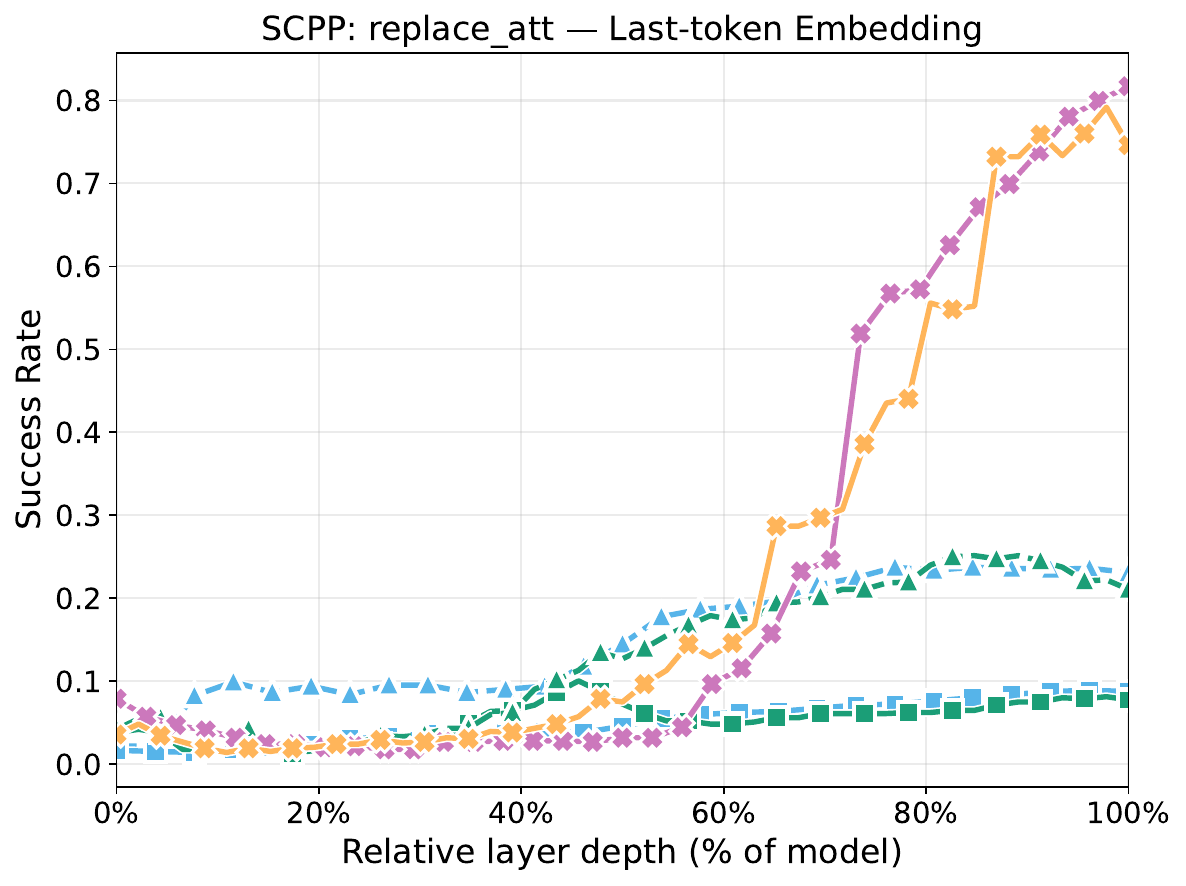}
    \end{subfigure}
    \hfill
    \begin{subfigure}{0.32\textwidth}
        \centering
        \includegraphics[width=\linewidth]{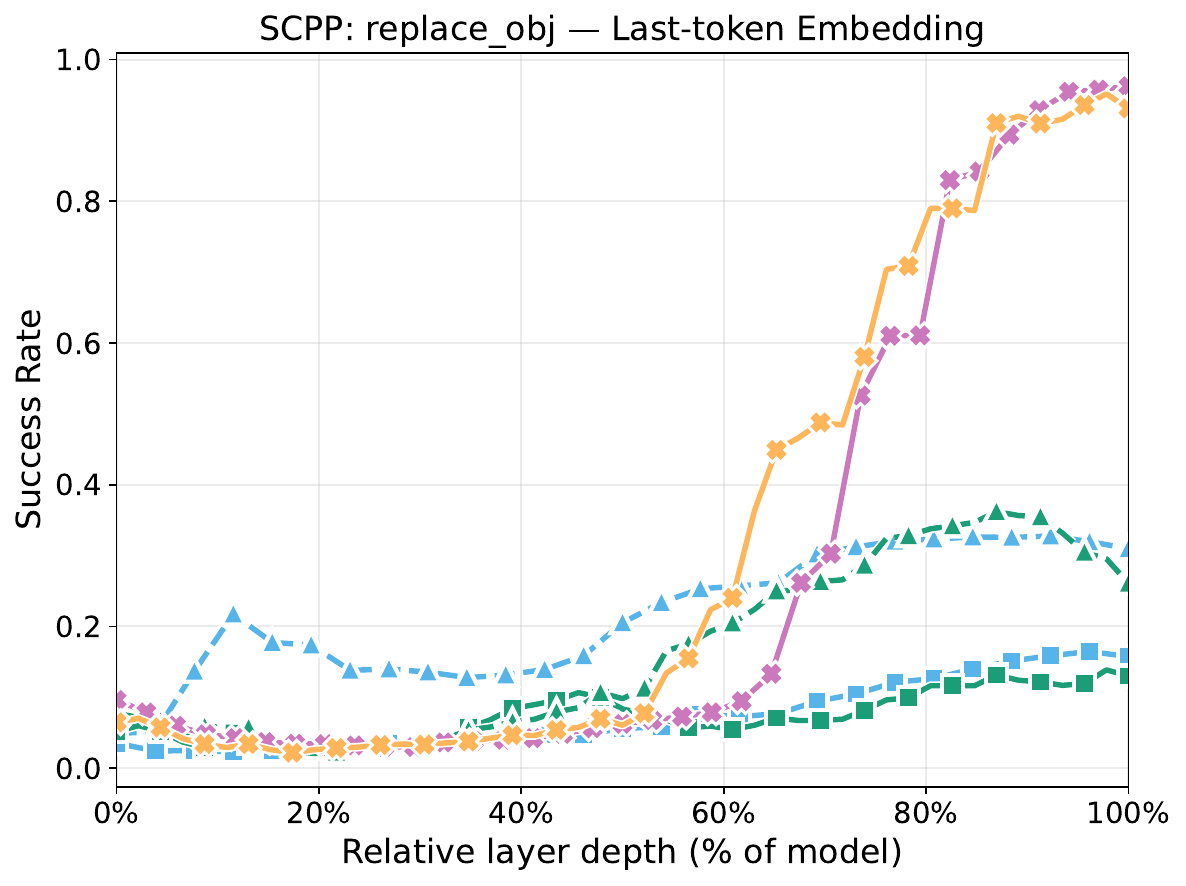}
    \end{subfigure}
    \begin{subfigure}{0.32\textwidth}
        \centering
        \includegraphics[width=\linewidth]{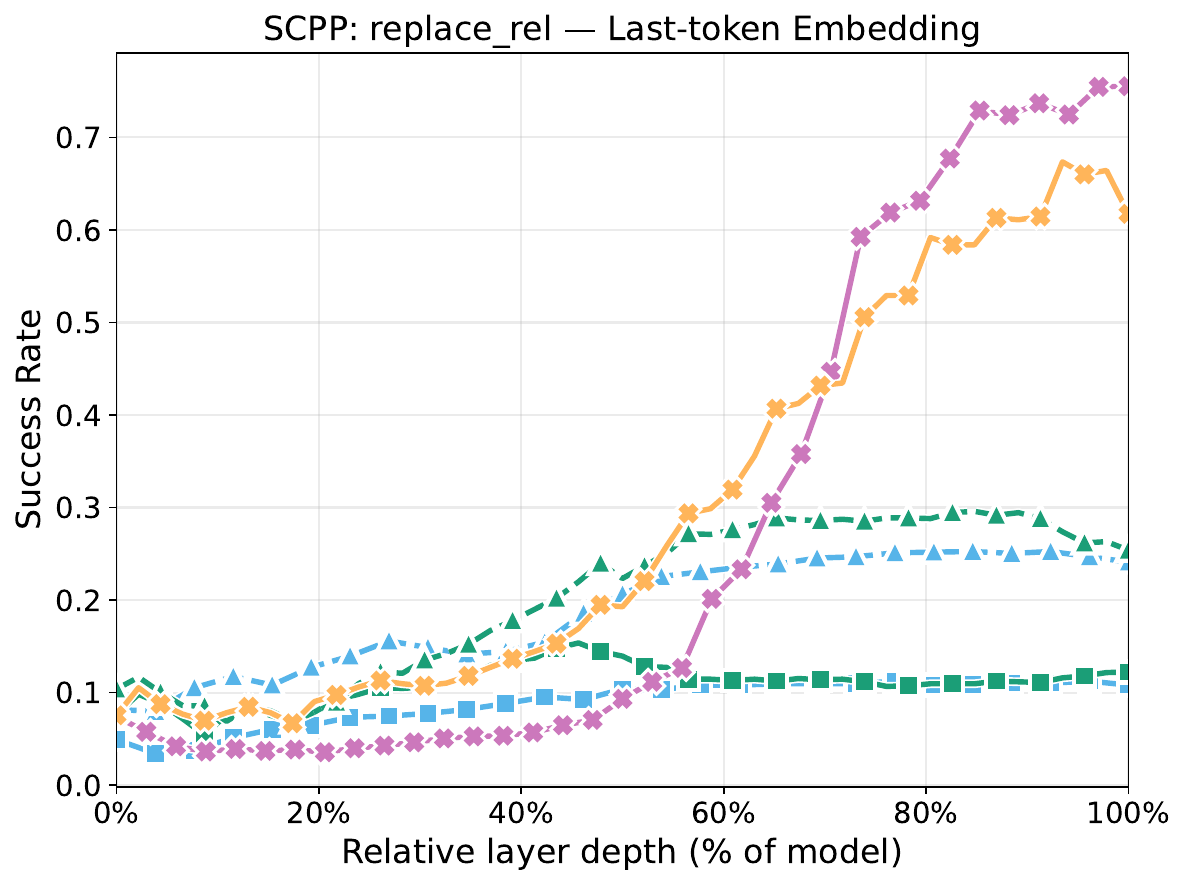}
    \end{subfigure}

    \begin{subfigure}{0.32\textwidth}
        \centering
        \includegraphics[width=\linewidth]{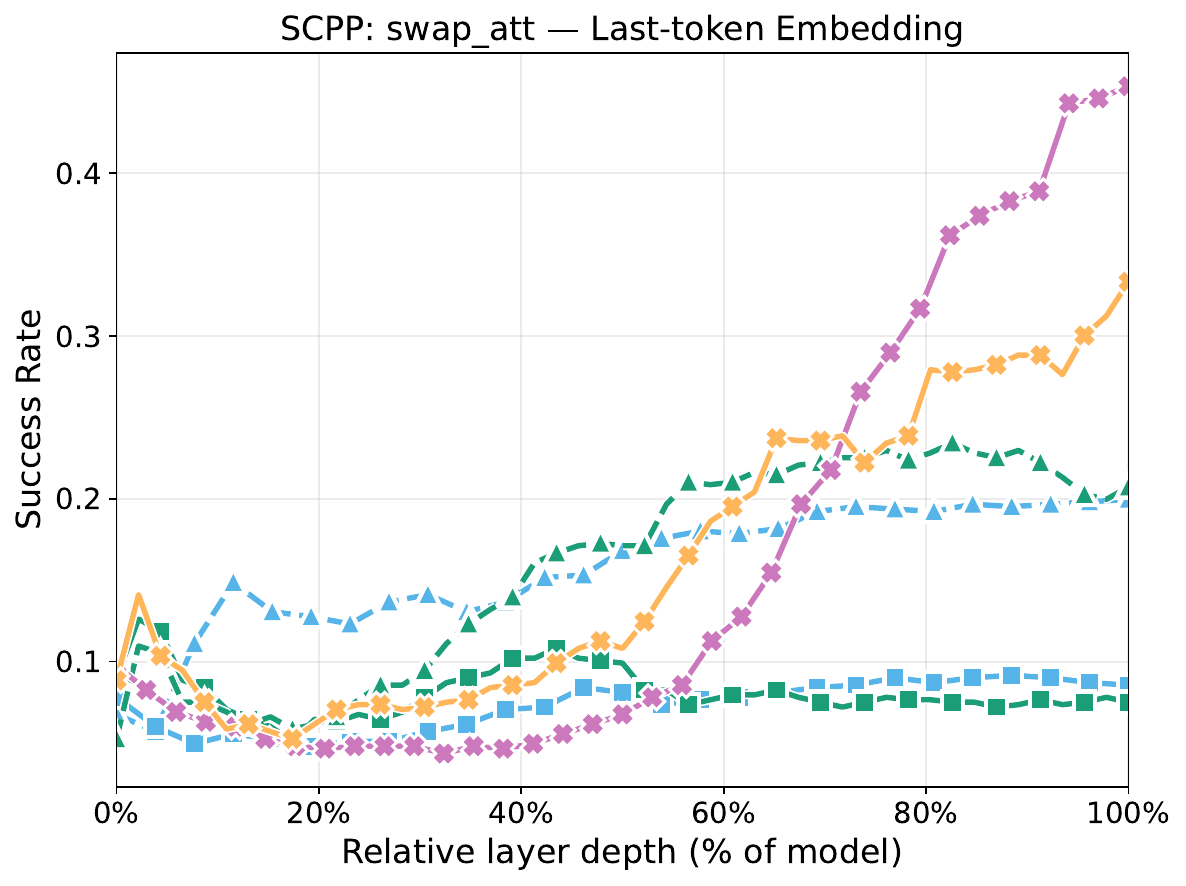}
    \end{subfigure}
    \begin{subfigure}{0.32\textwidth}
        \centering
        \includegraphics[width=\linewidth]{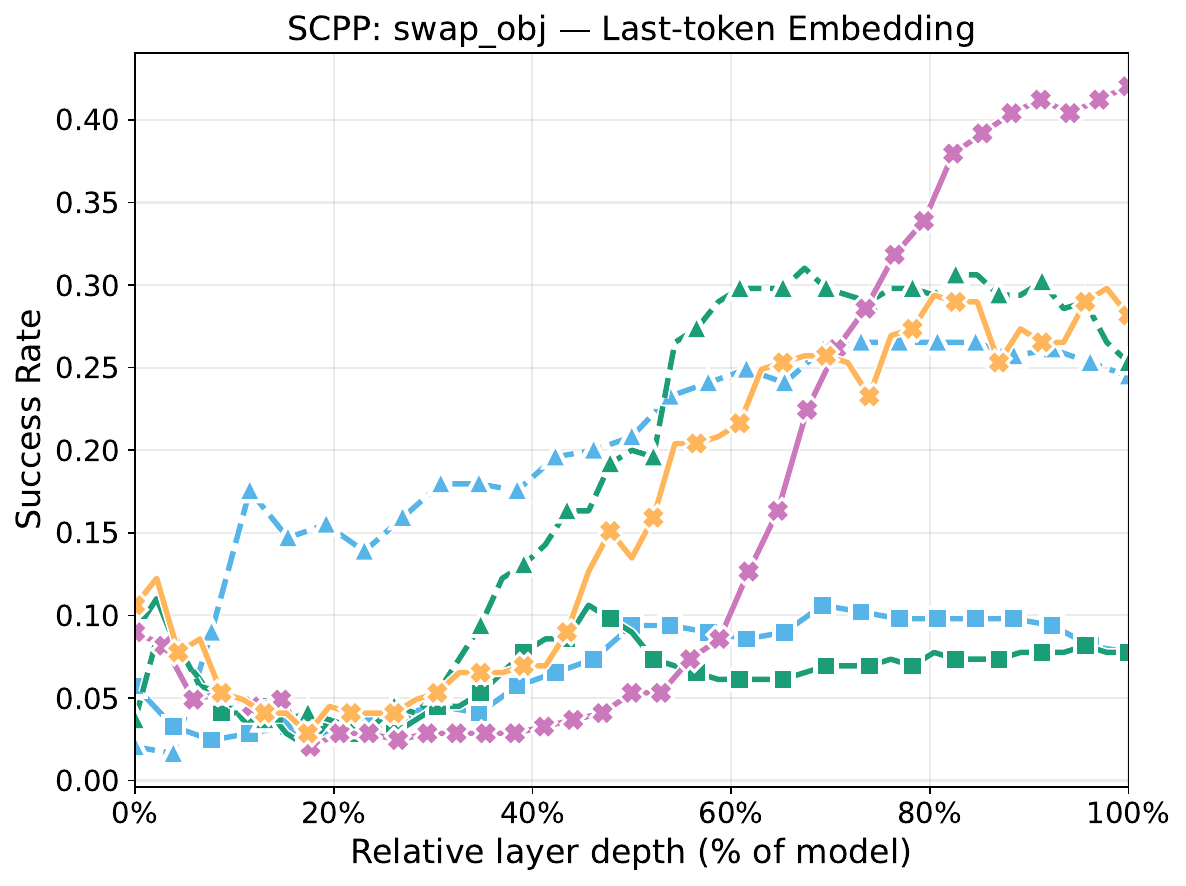}
    \end{subfigure}
    \begin{subfigure}{0.9\textwidth}
        \centering
        \includegraphics[width=\linewidth]{mteb_legend.pdf}
    \end{subfigure}

    \caption{Results for SCPP last token embeddings}
    \label{fig:appendix_scpp_lasttoken}
\end{figure*}

\section{Results for Lexical Influence Test on SCPP (subtasks)}
\label{sec:appendix_scpp_results}

Figure~\ref{fig:appendix_scpp_average},~\ref{fig:appendix_scpp_lasttoken} shows performance for the selected averaged and last token embeddings, respectively, for models referenced in \S~\ref{sec:experiment_setup}. 

\section{Effect on Lexical Influence Test on Prompt Variation}
\label{sec:appendix_prompt_variation}

\begin{figure*}[!t]
    \centering
    \begin{subfigure}{0.48\textwidth}
        \centering
        \includegraphics[width=\linewidth]{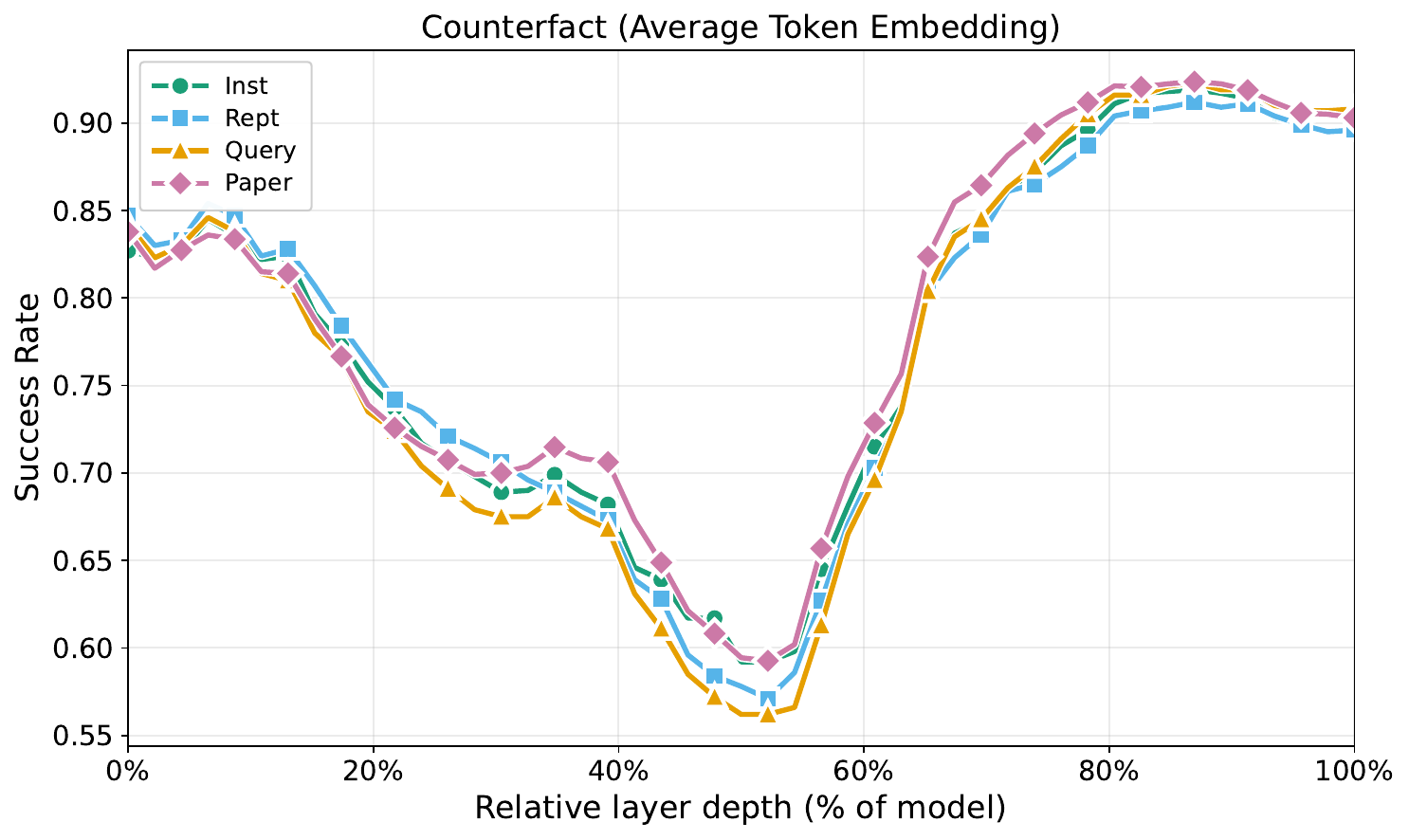}
    \end{subfigure}
    \hfill
    \begin{subfigure}{0.48\textwidth}
        \centering
        \includegraphics[width=\linewidth]{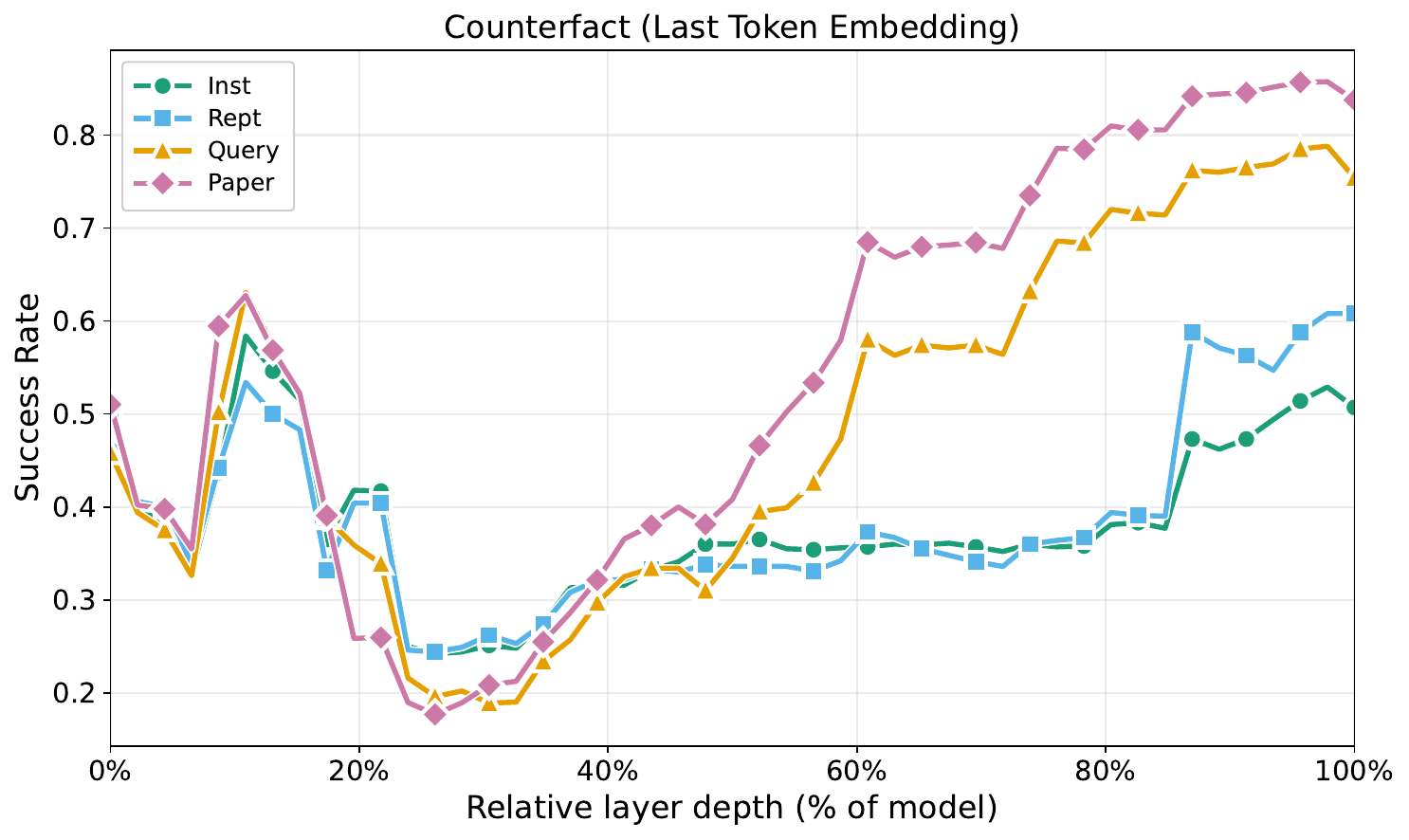}
    \end{subfigure}

    \caption{Results of prompt variations on the lexical influence test using the Gemma-12B-IT model on the CounterFact dataset.}
    \label{fig:prompt_variation_gemma}
\end{figure*}

To gauge the impact of baseline prompt variation on the instruction-tuned model, we evaluate the 
Gemma-3-12b-it on the CounterFact model under multiple instruction prompts. The list of prompts evaluated includes the following: (”query: ” (query), ”Represent this sentence for
searching relevant passages: ” (Rept), ”Instruct: Represent this text for retrieval. Query: ” (Inst), (Paper) contains no instruction other than the base structure for the instruction models.

The results are shown in Figure~\ref{fig:prompt_variation_gemma}. The results indicate that the overall pattern is preserved across prompt variations. To measure the correlation between the resulting curves, we compute the pairwise average distance correlation (dCor).
For mean-pooled embeddings, both the pattern
and performance are nearly unchanged, with dCor = 0.993. For last-token embeddings, variation is
larger in the second half of the network, where the model moves closer to its generation objective, but
the overall trend remains the same; the dCor is 0.78 in this case.

\section{Lexical and Semantic Probes}
\label{sec:appendix_lex_sem_probe}
\subsection{Lexical probe}
\label{sec:appendix_lex_probe}
\textbf{Data} We train the probe on raw WikiText (WikiText-103-raw). We tokenize the corpus with the same tokenizer as the base model, discard empty/whitespace-only lines, and concatenate text segments with an end-of-sequence boundary token to avoid creating spurious cross-document contexts. From this token stream, we construct fixed-length sequences of $L=50$ tokens (each sequence is used both as input and as next-token supervision with a standard causal language-modeling shift).

To mitigate degenerate solutions that rely disproportionately on high-frequency function words (e.g., stop words) rather than meaningful contextual cues, we promote diversity in the training signal by sampling sequences from a shuffled stream of documents (and, in the streaming setting, shuffling with a buffer) before chunking into length-$L$ windows. We additionally cap the number of extracted sequences per split (typically $3{,}000$) to control compute while maintaining broad topical coverage. Validation and test data are preprocessed identically but are never used for sampling or shuffling decisions.

\textbf{Training and Hyperparameters}
\label{sec:appendix_probe_train_hyperparams}

We train one probe independently for each selected layer while keeping the backbone language model frozen. For a given layer $l$, the probe applies a learned affine transformation to the layer's hidden states,
\[
    z_{l,t} = W_l H^M_l(x)_t + b_l,
\]
where $W_l$ is initialized to the identity and $b_l$ is initialized to zero. The transformed features are then $\ell_2$-normalized and scored against a row-normalized, frozen copy of the model's \emph{input embedding matrix}. Thus, the probe acts as a cosine-similarity classifier over the input vocabulary. We additionally learn a bounded per-layer logit scale to keep optimization stable.

We train the probe by minimizing the token-level cross-entropy loss $\mathcal{L}^{(l)}_{\mathrm{CE}}$ between the probe's scores and the ground-truth token IDs, ignoring padded positions. For causal models, we use the EOS token as the pad token. We also apply a weak $\ell_2$ penalty that encourages $W_l$ to remain close to the identity. This regularizer primarily stabilizes training and provides a mild preference for identity-like mappings, discouraging large remappings that can confound probe-based interpretation~\citep{HewittL19}. Consequently, probe accuracy reflects the linear accessibility of token identity in a space aligned with the model's input embedding geometry, which we use as an operational reference for surface form.

The resulting objective at layer $l$ is
\begin{equation}
   \mathcal{L}_{l}
  = \mathcal{L}_{\mathrm{CE},l} + \lambda \|W_l - I\|_F^2 ,
\end{equation}
where $\lambda > 0$ is the identity-regularization coefficient. We optimize the probe parameters with AdamW~\citep{loshchilov2017decoupled}, using learning rate $10^{-3}$, weight decay $10^{-4}$, identity regularization $\lambda = 10^{-6}$, and gradient-norm clipping to stabilize optimization.

\section{Semantic Probe}
\label{sec:appendix_semantic_probe}
Following the findings in Figure~\ref{fig:lexical_failures}, we use the average token (mean-pooled) for pretrained and instruction-tuned models and the last-token representation for embedding models. 

We evaluate on a suite of English tasks from the Massive Text Embedding Benchmark (MTEB) and cache each task locally to ensure deterministic reruns. We perform subsampling for the few \emph{massive} cases to keep runtime bounded. For \texttt{MedRxivClusteringS2S.v2} (37{,}500 test instances), we uniformly sample 10{,}000 test points without replacement using a fixed seed and cache the resulting subset. For extremely large reranking data such as \texttt{MindSmallReranking}, we cap the cached data to 10{,}000 queries and 200{,}000 corpus documents; we then filter \texttt{relevant\_docs} and \texttt{top\_ranked} to remove any entries that reference dropped query or document IDs, ensuring internal consistency. The datasets used and their categorization is provided in Table~\ref{tab:appendix_mteb_datasets}.

\subsection{Results MTEB}
\label{sec:appendix_mteb_results}
\begin{figure*}[!t]

    \centering
    \begin{subfigure}{0.48\textwidth}
        \centering
        \includegraphics[width=\linewidth]{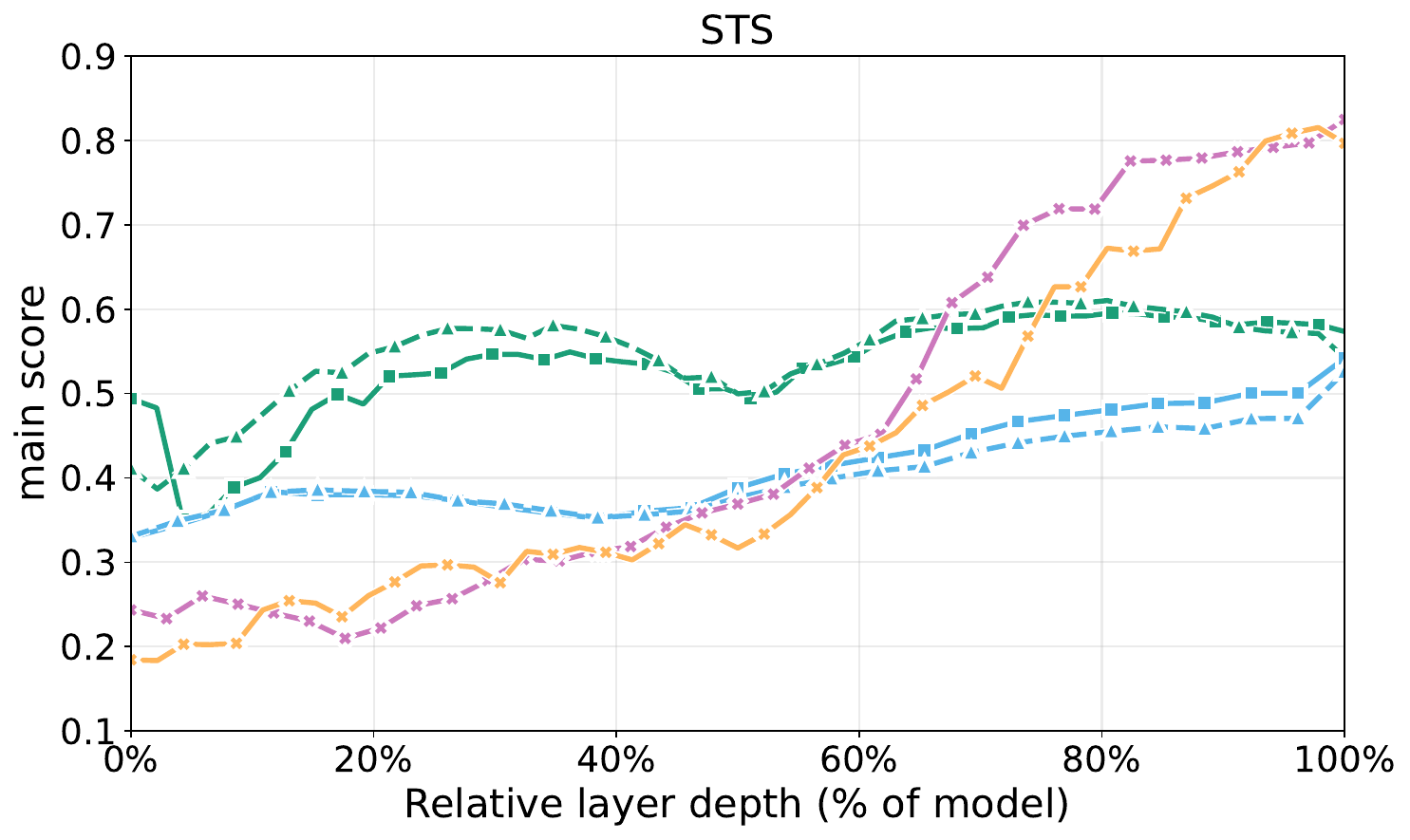}
    \end{subfigure}
    \hfill
    \begin{subfigure}{0.48\textwidth}
        \centering
        \includegraphics[width=\linewidth]{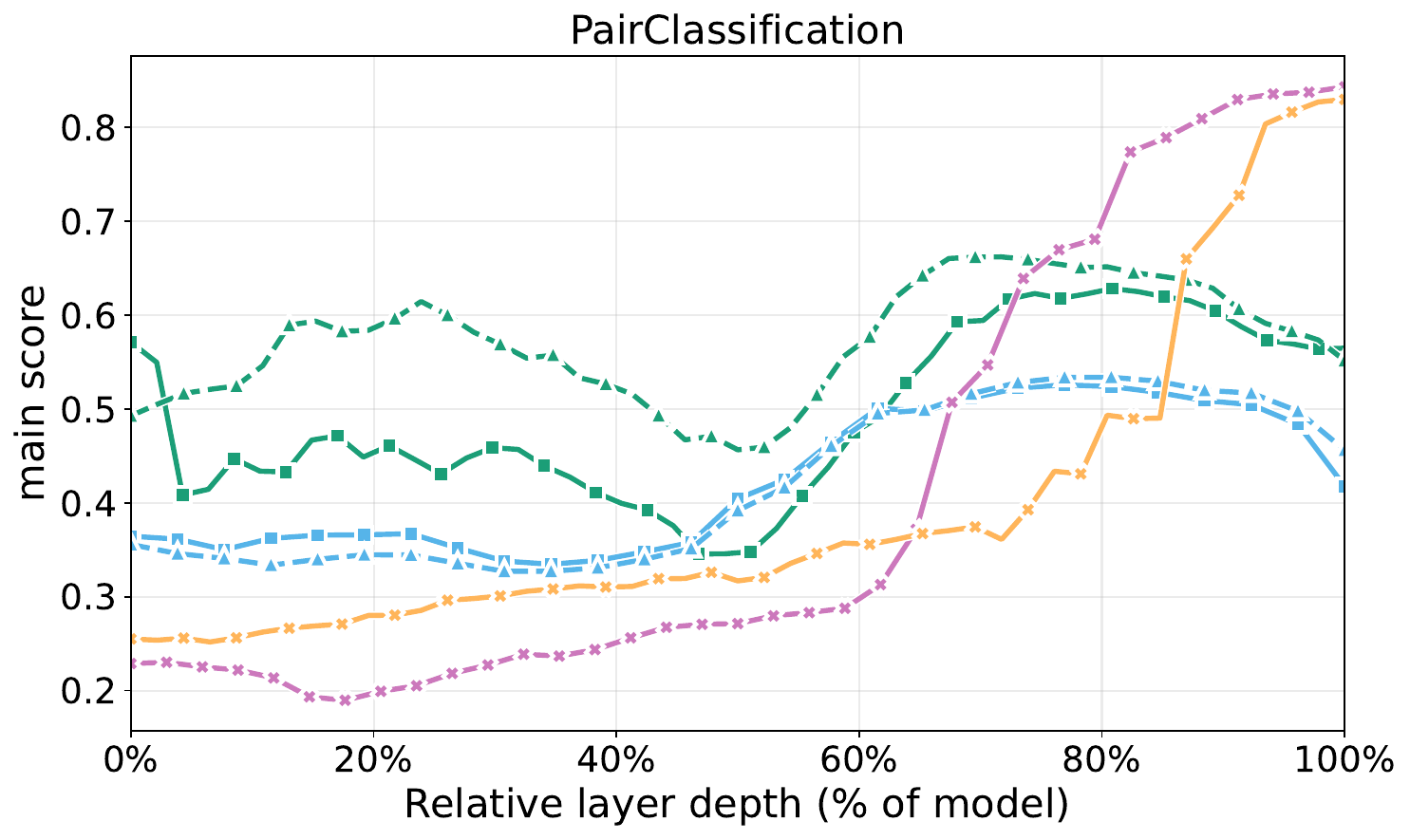}
    \end{subfigure}
    \begin{subfigure}{0.48\textwidth}
        \centering
        \includegraphics[width=\linewidth]{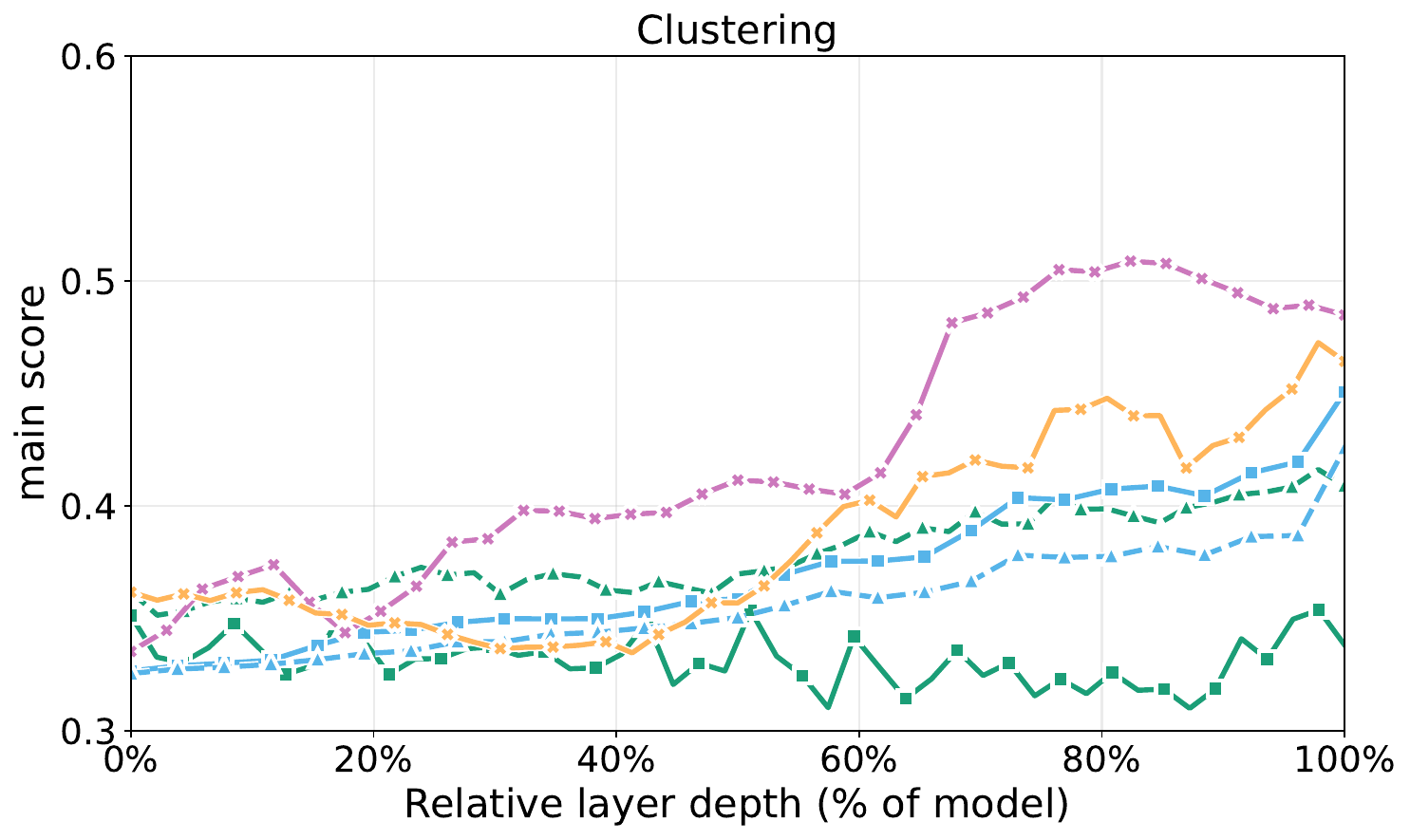}
    \end{subfigure}
    \begin{subfigure}{0.98\textwidth}
        \centering
        \includegraphics[width=\linewidth]{mteb_legend.pdf}
    \end{subfigure}

    \caption{Performance of the models on MTEB tasks. sts (a), pairwise classification (b), and clustering (c).}
    \label{fig:appendix_mteb_sts_summ}
\end{figure*}

Figure~\ref{fig:appendix_mteb_sts_summ} shows the results for STS, Pairwise Classification, and Clustering. Pairwise classification and sts follow the same geometric pattern discussed in the main text, while clustering diverges from the pattern. We hypothesize that this issue is related to the data and how evaluation is performed for this task; we leave this as future work.

\subsection{Word Sense Disambiguation}
\label{sed:appendix_wsd}
\begin{figure}[!t]
    \centering
    \includegraphics[width=0.5\linewidth]{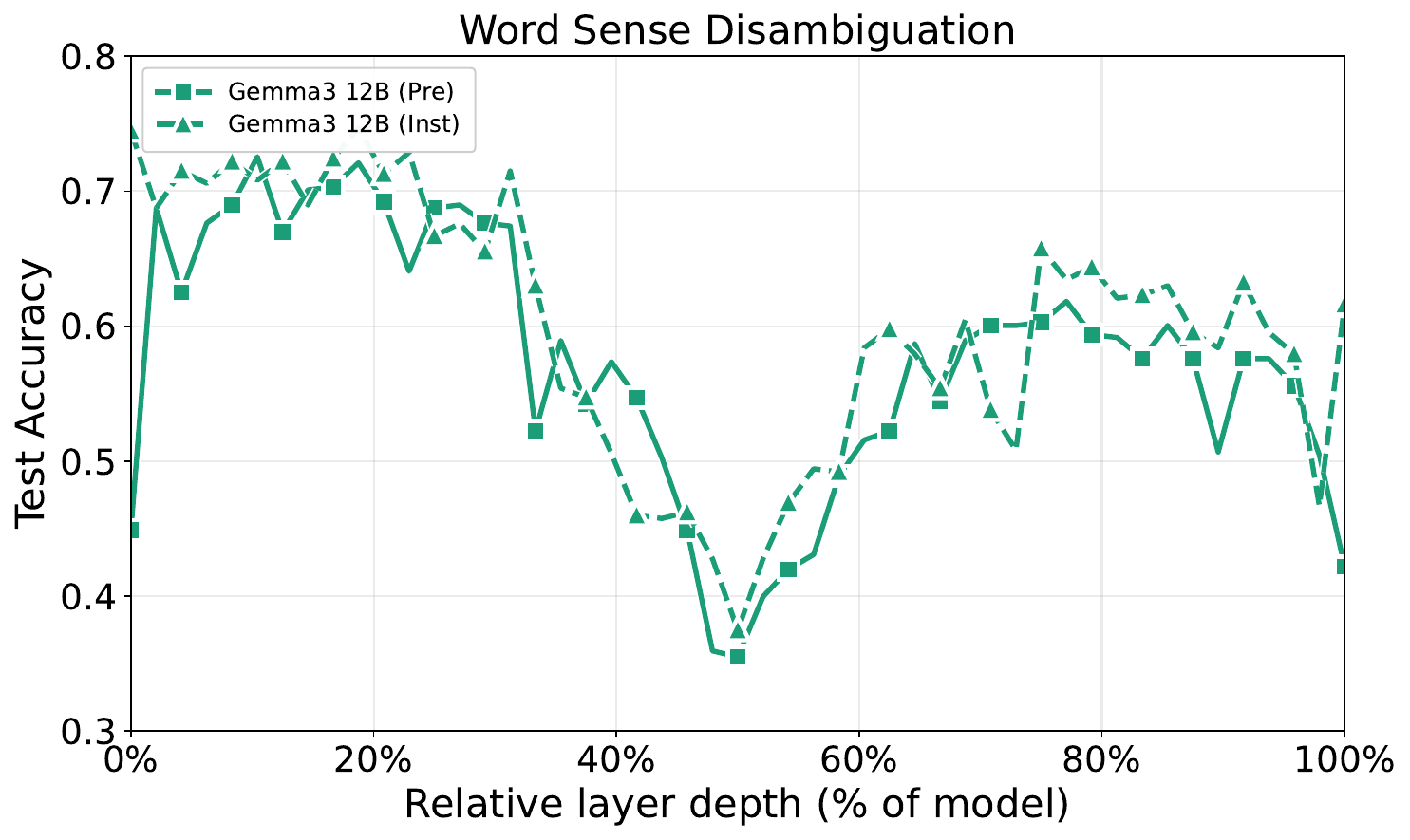}
    \caption{Results for WSD on Gemma-12B-IT  and Gemma-12B-PT model.}
    \label{fig:wsd}
\end{figure}

To validate the mid-depth degradation on token-level semantic tasks, we evaluate word sense disambiguation (WSD) on the SemCor corpus. We use token-level sense labels and restrict the prediction space to the 1,000 most frequent senses in the training set. For each layer, we train a linear probe and evaluate its performance. Figure~\ref{fig:wsd} shows a similar geometric pattern: performance degrades at intermediate depths and partially recovers in the latter half of the network. However, the later layers do not outperform the earlier ones. This is expected, as token representations in higher layers tend to become increasingly aligned with sequence-level semantics rather than preserving fine-grained token-specific information.

\begin{table}[t]
\vspace{-2mm}
\caption{MTEB task categories and datasets used.}
\label{tab:mteb_datasets_used}
\centering
\small
\begin{tabular}{p{0.28\linewidth} p{0.68\linewidth}}
\toprule
\textbf{Category} & \textbf{Datasets} \\
\midrule
Classification &
Banking77Classification; ToxicConversationsClassification; MTOPDomainClassification \\
Pair Classification &
ArguAna; SprintDuplicateQuestions; TwitterSemEval2015 \\
Reranking &
MindSmallReranking; AskUbuntuDupQuestions \\
Retrieval &
HotpotQAHardNegatives; TRECCOVID; SCIDOCS \\
Semantic Textual Similarity (STS) &
STS22.v2; STSBenchmark; SICK-R; BIOSSES \\
\bottomrule
\end{tabular}
\label{tab:appendix_mteb_datasets}
\end{table}

\section{Input/Prompt Entropy}
\label{sec:appendix_entropy}
Extracted from \citep{SkeanAZPNLS25}. For an input (prompt) $p$ with $T_p$ tokens, let $Z_\ell(p)\in\mathbb{R}^{T_p\times d}$ be the matrix of layer-$\ell$ token embeddings (rows correspond to token positions).
Define the Gram matrix
\[
K_\ell(p)=Z_\ell(p)\,Z_\ell(p)^{\top},
\qquad
\tilde{K}_\ell(p)=\frac{K_\ell(p)}{\operatorname{tr}(K_\ell(p))}.
\]
Let $\{\mu_i\}_{i=1}^{T_p}$ be the eigenvalues of $\tilde{K}_\ell(p)$ (so $\sum_i \mu_i=1$). The input (prompt) entropy is the matrix-based Rényi entropy
\[
H_{\mathrm{in}}^{(\ell)}(p)
=
\frac{1}{1-\alpha}\log\!\left(\sum_{i=1}^{T_p}\mu_i^{\alpha}\right),
\qquad \alpha>0,\ \alpha\neq 1,
\]
with the $\alpha\to 1$ limit (von Neumann/Shannon)
\[
H_{\mathrm{in}}^{(\ell)}(p)
=
-\sum_{i=1}^{T_p}\mu_i\log \mu_i.
\]

\section{Summarization}

\subsection{Details of Metrics}
\label{sec:summarization_details_method}

\textbf{BERTScore.}
BERTScore leverages contextual embeddings from a pre-trained Transformer (e.g., BERT) to compute semantic similarity between a candidate summary and a reference summary. Unlike $n$-gram metrics such as ROUGE, it captures synonymy and paraphrasing by computing cosine similarity between contextual token embeddings with greedy matching.

Given a reference summary $x=\langle x_1,\dots,x_{T_x}\rangle$ and a candidate summary $\hat{x}=\langle \hat{x}_1,\dots,\hat{x}_{T_{\hat{x}}}\rangle$, let $\mathbf{v}_i\in\mathbb{R}^d$ and $\hat{\mathbf{v}}_j\in\mathbb{R}^d$ denote their contextual token embeddings. Define the cosine similarity
\[
s_{ij}=\frac{\mathbf{v}_i^\top \hat{\mathbf{v}}_j}{\|\mathbf{v}_i\|_2\,\|\hat{\mathbf{v}}_j\|_2}.
\]
Then the recall ($R_{\text{BERT}}$), precision ($P_{\text{BERT}}$), and F1 ($F_{\text{BERT}}$) are
\begin{equation}
R_{\text{BERT}}=\frac{1}{T_x}\sum_{i=1}^{T_x}\max_{1\le j\le T_{\hat{x}}} s_{ij},
\end{equation}
\begin{equation}
P_{\text{BERT}}=\frac{1}{T_{\hat{x}}}\sum_{j=1}^{T_{\hat{x}}}\max_{1\le i\le T_x} s_{ij},
\end{equation}
\begin{equation}
F_{\text{BERT}}=2\cdot\frac{P_{\text{BERT}}\,R_{\text{BERT}}}{P_{\text{BERT}}+R_{\text{BERT}}}.
\end{equation}

We report the F1 score ($F_{\text{BERT}}$). We follow the official BERTScore configuration and use the default DeBERTa layer (40), which is tuned for WMT16 human-correlation.

\textbf{BARTScore.} 
BARTScore treats the evaluation of generated text as a sequence-to-sequence generation task. It utilizes the log-probability of the target sequence (the candidate summary) conditioned on a source sequence (either the reference or the original document) using a pre-trained BART model. This allows the metric to capture aspects like faithfulness and informativeness more effectively than surface-level matching.

The score is defined as the average log-likelihood of the target sequence $y$ given the source sequence $x$:

\begin{equation}
\text{BARTScore} = \frac{1}{m} \sum_{t=1}^{m} \log P(y_t | y_{<t}, x, \theta)
\end{equation}

where $\theta$ represents the parameters of the pre-trained BART model, and $m$ is the length of the target sequence.

\subsection{Data and Models}
\label{sec:summarization_data_evaluation}

\textbf{Data.} We use the standard extreme summarization\citep{narayan2018don} benchmark. We utilize randomly sampled 500 samples from the dataset. To probe lexical sensitivity, we augment each example with a highly lexicalized non-summary distractor generated by an LLM(gpt-5.2-mini, Temperature = 0.7). The prompt used for the generation of the distractor is as follows:

\begin{lstlisting}[language=Python, caption=Prompt for Lexical Distractor Non-Summary Generation,numbers=none]
prompt = f"""
    You are generating adversarial evaluation data to test lexical bias in summarization systems.
    
    Source Document:
    {document}
    
    Target Length: {target_length} characters (+/-5%)
    
    TASK:
    Generate a text that has HIGH lexical overlap with the source document but NO semantic overlap with it.
    
    CRITICAL DEFINITION:
    "NO semantic overlap" means that the text must not convey the topic, purpose, claims, events, or conclusions of the source in any form. 
    A reader should be unable to infer what the source document is about from your text.
    
    CONSTRAINTS (hard requirements):
    
    1. Lexical Overlap:
    - Reuse distinctive words, phrases, and named entities from the source.
    - Do not introduce new terminology.
    - Try to use all words from the source.
    
    2. Semantic Orthogonality:
    - Do NOT describe, summarize, or restate any part of the source's meaning.
    - Do NOT preserve relations, causality, chronology, or argumentation.
    - The text must be about something else entirely, despite using the same words.
    
    3. Length:
    - Approximately {target_length} characters.
    
    VALIDATION CHECK (internal):
    If the output allows a reader to identify the subject or purpose of the source document, it fails.
    
    OUTPUT FORMAT:
    Return a JSON object with exactly one field:
    
    {{
    "lexical_non_summary": "<text>"
    }}
\end{lstlisting}
\textbf{Models.}
For BERTScore, DeBERTa-XLarge~\cite{he2021deberta} is used. For BARTScore, we utilize the \textit{bart-large}~\cite{lewis2019bart} model fine-tuned on the CNN/Daily Mail summarization dataset.

\section{Model Editing}
\label{sec:appendix_model_editing}
\subsection{Method and dataset.}
\paragraph{AlphaEdit (null-space constrained editing).}
AlphaEdit is a simple refinement that can be applied on top of a base locate--then--edit method (e.g., ROME/MEMIT). 
Suppose the base editor proposes an update $\Delta_{\text{base}}$ to a weight matrix $W$ to enforce the desired edit. 
AlphaEdit additionally builds a \emph{preservation set} of inputs whose behavior should remain unchanged, and extracts their corresponding key vectors $\{k_i\}_{i=1}^m$ at the edited layer. 
Let $K_0 \in \mathbb{R}^{d \times m}$ be the matrix with columns $k_i$. 
AlphaEdit then projects the base update onto the subspace orthogonal to $\mathrm{span}(K_0)$:
\[
\Delta_{\alpha} \;=\; \Pi_{\perp}\,\Delta_{\text{base}},
\qquad
\text{where }\Pi_{\perp}\text{ projects onto } \mathrm{span}(K_0)^\perp .
\]
This guarantees
\[
\Delta_{\alpha} K_0 = 0,
\]
so the refined update leaves the model's mapping on the preservation keys unchanged, while retaining the base editor's effect on the target edit.

\textbf{CounterFact dataset.} CounterFact’s distractor prompts are already partially lexically aligned with the anchor/edit prompt. To more directly quantify the effect of model editing under controlled lexical overlap, we augment CounterFact with two additional prompt sets: (\emph{i}) \emph{lexically similar} prompts, obtained by substituting the subject entity in the original anchor/edit template with distractor entities; and (\emph{ii}) \emph{lexically dissimilar} prompts, which preserve the distractor’s semantics while substantially rewriting its surface form. We generate the lexically dissimilar prompts using an LLM (gpt-5.2-mini) with the following prompt:

\begin{lstlisting}[language=Python, caption=Data generation prompt for lexicalLy similar distractor/locality,numbers=none]
prompt = f"""
You are a language expert tasked with rewriting prompts.

Input fields:
- original_locality_prompt: the prompt whose meaning must be preserved exactly
- edit_prompt: a reference prompt whose wording you must avoid copying

Task:
Produce THREE rewritten variants of original_locality_prompt.

STRICT meaning preservation (each variant):
1) Ask for the exact same fact as original_locality_prompt: same entity name(s) + same attribute/relation.
2) Copy entity name(s) EXACTLY as in original_locality_prompt (same spelling/casing). Do not replace entities with descriptions.
3) Do NOT add/remove qualifiers (time, certainty, official/current, "capital", "city of", etc.). No hints, no assumptions.

Cloze formatting (each variant):
4) Each prompt must be a PREFIX such that the correct answer should be the next text generated immediately after the final space.
5) End with exactly ONE trailing ASCII space and NO trailing punctuation (no ., ?, !, :, ;, ,).
6) Do NOT end with the entity name(s).
7) No interrogatives: do NOT use WH-words (what/which/where/who/when/how). Use declarative stems only.
8) Validity check: appending the correct answer immediately after the final space must yield a complete grammatical result without adding any extra words.

Anti-copy / maximize lexical difference from edit_prompt (each variant):
9) Hard ban: Do NOT reuse any contiguous 2+ word sequence found in edit_prompt (exception: entity tokens).
10) Content-word taboo: Avoid reusing ANY non-entity content word from edit_prompt (verbs/nouns/adjectives/adverbs).
    - Allowed function-word set (may repeat): {the, a, an, of, to, in, on, at, from, for, with, by, is, are, was, were, be, been, being, as}
    - If a relation word from original_locality_prompt also appears in edit_prompt, you should paraphrase it using a synonym NOT present in edit_prompt, while preserving meaning.
11) Target: minimize overlap of non-entity content words with edit_prompt to 0 whenever possible.

MANDATORY structural diversity (across the 3 variants):
Generate exactly one from each frame:

A) ENTITY-SUBJECT frame:
   "<ENTITY> <connector> "
B) POSSESSIVE / ownership frame:
   "<ENTITY>'s <attribute phrase> <connector>"  OR  "<attribute phrase> of <ENTITY> <connector>"
C) ATTRIBUTE-SUBJECT frame:
   "The <attribute phrase> for <ENTITY> <connector>"

Additional diversity constraints:
12) No shared opening 3 tokens across variants.
13) Use three different connector tails (the last 1 to 4 words before the final space must differ across variants).
14) Across the three outputs, do not reuse the same key content words (besides the entity and unavoidable attribute noun). Prefer different synonym sets.

Search-and-select requirement (important):
15) For EACH frame (A/B/C), internally draft at least 5 candidate rewrites, then choose the best one that:
    - satisfies all constraints,
    - has the fewest non-entity content words overlapping with edit_prompt,
    - and differs most from the other selected variants.

Forbidden:
- Do not include the literal strings "original_locality_prompt" or "edit_prompt".
- Output JSON only; no extra text.

Output JSON only (exactly these keys, no extras):
{"rewritten_prompts":["... ","... ","... "]}
"""
\end{lstlisting}

We perform editing on 2,000 sequential edit samples using the Llama-3-8B model, which is directly supported by the AlphaEdit codebase and evaluated with the default base settings.

\subsection{Metrics}
Let $M_0$ be the base model and $M_1$ the edited model. For an input $x$, let $\tau_x$ denote the answer position.
Let $p_x^{M}(v)=P_M(v\mid x,T_x)$ be the next-token distribution over the vocabulary $\mathcal V$.

\paragraph{Merged top-$K$ support.}
For $K\in\{20,50,100\}$, define
\[
S_K(x)=\mathrm{Top-K}\!\bigl(p_x^{M_0}\bigr)\ \cup\ \mathrm{Top-K}\!\bigl(p_x^{M_1}\bigr),
\qquad
Z_x^{M}(K)=\sum_{u\in S_K(x)} p_x^{M}(u),
\]
and the renormalized distributions on the merged support
\[
\tilde p_{x,K}^{M}(v)=\frac{p_x^{M}(v)}{\max\{Z_x^{M}(K),\varepsilon_{\text{stab}}\}},
\qquad v\in S_K(x),
\]
with $\varepsilon_{\text{stab}}=10^{-12}$. All @K divergence metrics are computed on
$(\tilde p_{x,K}^{M_0},\tilde p_{x,K}^{M_1})$.

\paragraph{Jensen--Shannon divergence (JSD@K).}
Let $m_{x,K}=\tfrac{1}{2}\bigl(\tilde p_{x,K}^{M_0}+\tilde p_{x,K}^{M_1}\bigr)$. Then
\[
\mathrm{JSD}_K(x)
=
\frac{1}{2}\mathrm{KL}\!\left(\tilde p_{x,K}^{M_0}\,\|\,m_{x,K}\right)
+
\frac{1}{2}\mathrm{KL}\!\left(\tilde p_{x,K}^{M_1}\,\|\,m_{x,K}\right),
\]
where
\[
\mathrm{KL}(r\|s)=\sum_{v\in S_K(x)} r(v)\log\frac{\max\{r(v),\varepsilon_{\text{stab}}\}}{\max\{s(v),\varepsilon_{\text{stab}}\}}.
\]

\paragraph{Total variation distance (TV@K).}
\[
\mathrm{TV}_K(x)
=
\frac{1}{2}\sum_{v\in S_K(x)}
\left|\tilde p_{x,K}^{M_0}(v)-\tilde p_{x,K}^{M_1}(v)\right|.
\]

\paragraph{Kendall's $\tau$ (rank stability on $S_K(x)$).}
Let $r_{x,K}^{M}(v)$ be the rank of token $v\in S_K(x)$ when sorting $p_x^{M}(v)$ in descending order.
Over unordered pairs $\{u,v\}\subset S_K(x)$, define
\[
a=r_{x,K}^{M_0}(u)-r_{x,K}^{M_0}(v),\qquad
b=r_{x,K}^{M_1}(u)-r_{x,K}^{M_1}(v).
\]
Pairs with $a=0$ or $b=0$ (ties in either ranking) are skipped. Let $C$ and $D$ be the number of remaining
concordant and discordant pairs, where concordant means $ab>0$ and discordant means $ab<0$. Then
\[
\tau_K(x)=
\begin{cases}
\frac{C-D}{C+D}, & C+D>0,\\[3pt]
0, & C+D=0.
\end{cases}
\]

\paragraph{Top-1 flip rate.}
\[
\mathrm{Flip}(x)=
\mathbf{1}\!\left[
\arg\max_{v\in\mathcal V} p_x^{M_0}(v)\neq
\arg\max_{v\in\mathcal V} p_x^{M_1}(v)
\right].
\]

\paragraph{Edited-Target Presence@K (leakage onto locality prompts).}
Let $\mathcal{T}\subset\mathcal V$ be the set of relation-specific edited target token ids, and let $\mathcal{L}$ be the
set of locality/distractor inputs. For $x\in\mathcal L$,
\[
\mathrm{ETP}_K(x)=
\mathbf{1}\!\left[
\mathrm{Top-K}\!\bigl(p_x^{M_1}\bigr)\cap \mathcal T \neq \emptyset
\right],
\qquad
ETP_K=\frac{1}{|\mathcal L|}\sum_{x\in\mathcal L}\mathrm{ETP}_K(x).
\]

\textbf{Results} Results for top-K $100$ are provided in Table~\ref{tab:appendix_model_editing}.
\begin{table}[t]
\centering
\small
\caption{Distribution-shift metrics on \emph{lexically similar and dissimilar} prompts (Top-K-100).}
\setlength{\tabcolsep}{7pt}
\renewcommand{\arraystretch}{1.15}
\begin{tabular}{lcc}
\toprule
\textbf{Metric} & \textbf{LS} & \textbf{LD} \\
\midrule
JSD                    & 0.294 & 0.246 \\
TV                     & 0.568 & 0.502 \\
Kendall $\tau$         & 0.001 & 0.039 \\
Edited-Target Presence & 0.142 & 0.072 \\
\midrule
Top-1 Flip Rate \textit{(Top-K-independent)} & 0.566 & 0.493 \\
\bottomrule
\end{tabular}
\label{tab:appendix_model_editing}
\end{table}

\section{Compute Resources}
\label{sec:compute}
All experiments are conducted on an NVIDIA H100 GPU with 96 GB of VRAM. Several experiments, including the semantic probing evaluations, require storing intermediate embeddings on disk; these runs used up to 512 GB of storage and 128 GB of system RAM.

\end{document}